\documentclass[letterpaper,twocolumn,10pt]{article}
\usepackage{usenix2019_v3}
\usepackage{lineno,comment}

\usepackage{pifont}
\usepackage{epsfig,epsf,url}
\usepackage{tabularx}
\usepackage{float}
\usepackage[linesnumbered,boxed,ruled,vlined]{algorithm2e}
\usepackage{algorithmic}
\usepackage{amsmath, amssymb}
\usepackage{mathtools}
\usepackage{amsthm}
\usepackage{rotating}
\usepackage{diagbox}
\usepackage{makecell}
\usepackage{multirow}
\usepackage{hhline}
\usepackage{booktabs}

\usepackage{paralist}
\newenvironment{icompact}{
  \begin{list}{$\bullet$}{
    \parsep 0.5pt plus 0.5pt
    \partopsep 0.5pt plus 0.5pt
    \topsep 0.5pt plus 1pt minus 0.5pt
    \itemsep 0.5pt plus 0.5pt
    \parskip 0pt plus 1pt
    \leftmargin 0.23in
    \labelsep 0.05in}
       }
  {\normalsize\end{list}}

\usepackage{epstopdf}
\long\def\comment#1{}
\usepackage{multirow}
\usepackage{lscape}
\usepackage{stmaryrd}
\usepackage{wrapfig}
\usepackage{hhline}
\usepackage{fancyvrb}
\usepackage{textcomp,booktabs}
\usepackage{color}
\usepackage{subcaption}
\usepackage{caption}
\usepackage{listings}
\newcommand{\eg}{{\it e.g.}}

\newcommand{\sys}{{\textsc{TideRL}}\xspace}
\newcommand{\rap}{$\mathrm{RA}^2\mathrm{P}$\xspace}
\newcommand{\boldrap}{$\mathbf{RA}^2\mathbf{P}$\xspace}

\newcount\hour \newcount\minute
\hour=\time  \divide \hour by 60
\minute=\time
\loop \ifnum \minute > 59 \advance \minute by -60 \repeat
\def\drafttime{\ifnum \hour<13 \number\hour:%
                      \ifnum \minute<10 0\fi
                      \number\minute
                      \ifnum \hour<12 \ AM\else \ PM\fi
         \else \advance \hour by -12 \number\hour:%
                      \ifnum \minute<10 0\fi
                      \number\minute \ PM\fi}

\usepackage{cleveref}
\crefname{section}{}{\S\S}
\crefdefaultlabelformat{\S#2#1#3}

\crefname{appendix}{Appendix}{Appendices}
\creflabelformat{appendix}{#2#1#3}
\crefname{subappendix}{Appendix}{Appendices}
\creflabelformat{subappendix}{#2#1#3}

\usepackage{hyperref}
\usepackage{balance}
\usepackage{tikz}

\makeatletter
\def\@fnsymbol#1{\ensuremath{\ifcase#1\or *\or \dagger\or \ddagger\or
   \mathsection\or \mathparagraph\or \|\or **\or \dagger\dagger\else\@ctrerr\fi}}
\makeatother
\begin{document}

\title{\Large \bf \sys: Boosting Agentic RL Goodput with Readiness-Aware Scheduling}

\author{
  \rm{Yanyu Ren$^{1*\dagger}$, Xizheng Wang$^{3*}$, Xiao Liu$^{1,2*}$,
        Bowen Lv$^{1\dagger}$, Hanchen Zhang$^{1\dagger}$, Shudan Zhang$^{1\dagger}$,} \\
  \rm{Hanyu Lai$^{1\dagger}$, Shuai Wang$^{3}$, Li Chen$^{3}$, Dan Li$^{1}$, Jie Tang$^{1}$} \\
  \textsuperscript{1}Tsinghua University \hspace{2ex}
  \textsuperscript{2}Z.AI \hspace{2ex}
  \textsuperscript{3}Zhongguancun Laboratory
}

\maketitle
\begingroup
\renewcommand{\thefootnote}{\fnsymbol{footnote}}
\footnotetext[1]{Equal contribution.}
\footnotetext[2]{Work done while these authors interned at Z.AI.}
\endgroup
\setlength{\textfloatsep}{0.5pt}% Remove \textfloatsep
\setlength{\abovecaptionskip}{0pt plus 1pt minus 1pt}
\setlength{\belowcaptionskip}{0pt plus 1pt minus 1pt}
\medmuskip=0mu\relax % Remove spaces around $\times$
\thinmuskip=0mu\relax % Remove spaces around $\times$
\thickmuskip=0mu\relax % Remove spaces around $\times$
\setlength{\abovedisplayskip}{0pt}
\setlength{\belowdisplayskip}{0pt}
\setlength{\abovedisplayshortskip}{0pt}
\setlength{\belowdisplayshortskip}{0pt}

\begin{abstract}
Reinforcement learning~(RL) for large language models is moving toward multi-turn agentic workloads, where rollout tasks repeatedly pause for external environments, resume with growing contexts, and finish at highly variable times. In this setting, RL training goodput, measured by training throughput, matters more than raw GPU occupancy: GPU waiting and repeated prefill recomputation are pure overhead. We present \sys, a readiness-aware elastic RL system with Continuous Task Batching, Resource-Aware Ref-Actor Pipelining, and Elastic Resource Scaling. CTB preserves useful rollout state, \rap selects between decoupled streaming and colocated aggregation from the ready backlog and arrival interval, and ERS moves ranks between rollout and training using the same readiness signals. Across text-only and multi-modal agentic workloads, \sys improves RL training goodput by up to 5.6$\times$ over synchronous baselines and over 33\% over asynchronous baselines, while reaching similar task performance. It also improves KV cache hit rate by 1.58$\times$, reduces per-step training time by up to 44.3\%, and cuts total waiting time by up to 77.6\%.
\end{abstract}

%!TEX root = ../main.tex
\section{Introduction}
\label{sec:intro}

Large Language Models (LLMs) are rapidly evolving from static text generators into autonomous agents capable of solving complex, real-world problems. In these agentic workloads, models do not merely produce a single-turn answer; instead, they iteratively interact with external, distributed environments, such as web browsers, smartphones, or computer OSes~\cite{webarena, osworld, androidlab}. This introduces a multi-turn execution flow. An agent generates an action, transmits it over the network to the environment, waits for the execution results, and resumes generation based on the newly returned observations. 
Because LLM outputs are non-deterministic, execution trajectories of the same task may differ drastically in turn count, context length, and wall-clock duration, producing workloads with extreme variance across both spatial and temporal dimensions.
% Due to the nondeterministic nature of LLM outputs, even the same agentic task may take a different number of turns to accomplish, leading to highly variable context lengths across different trajectories.

% \begin{figure}[tb]
%     \centering
%     % Placeholder for the problem-driven system workflow figure
%     \includegraphics[width=1.0\linewidth]{figs/system_workflow.png}
%     \caption{\sys brings benefit to agentic RL at three levels: rollout, training and overall scheduling.}
%     \label{fig:system_workflow}
% \end{figure}

\begin{figure}[tb]
    \centering
    \includegraphics[width=0.95\linewidth]{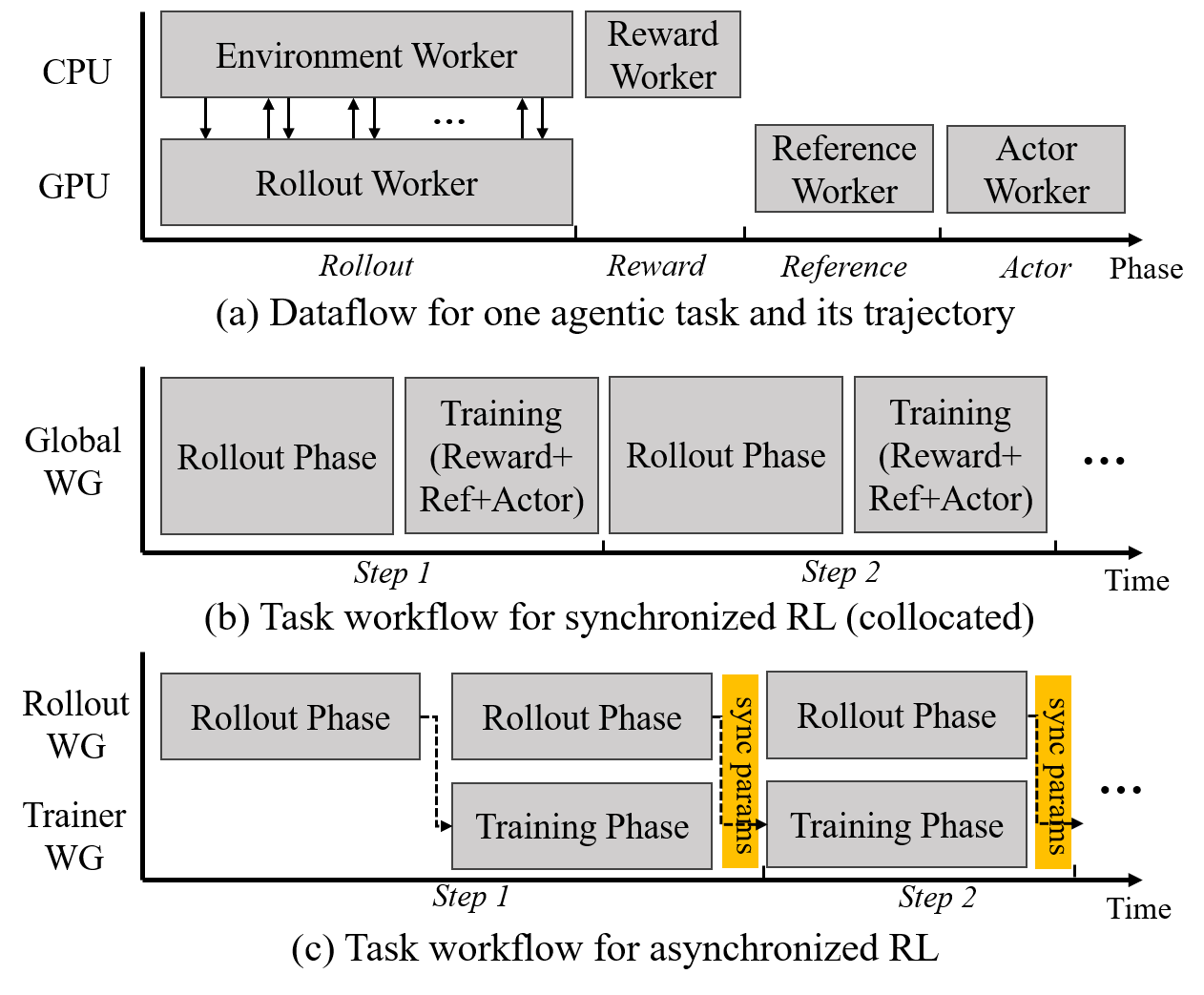}
    \caption{The dataflow and workflow of different RL strategies on agentic tasks.}
    \label{fig:concept_arch}
\end{figure}

To enhance these autonomous capabilities, researchers turn to reinforcement learning~(RL). While RL has shown immense success in single-turn reasoning tasks~\cite{openrlhf, grpo, ppo}, applying it to multi-turn agentic scenarios requires orchestrating massive, networked machine learning clusters. A typical RL workflow goes through four phases. In the \textit{rollout} phase, a batch of agentic tasks runs concurrently, requesting actions from rollout workers and observations from environment workers to generate interactive trajectories in multiple turns. Later, the RL system evaluates these trajectories in the \textit{reward} and \textit{reference} phase. Finally, the system uses the output from the reward and the reference to update the model weights in the \textit{actor} phase.

However, a small number of long-tail tasks in the rollout phase may block the pipeline and leave expensive GPU resources underutilized. For agentic RL, the right objective is goodput, which we measure through training throughput: the rollout'd tokens that reach the Trainer and advance an update counts. GPU waiting and repeated prefill recomputation are pure overhead. To maximize goodput, asynchronous systems~\cite{streamrl,areal} separate these components onto different GPUs, unlike synchronous systems like VeRL~\cite{verl} that force rollout-train barriers. On a subset of GPUs, \textit{rollout worker groups} perform continuous generation as producers. On the remaining GPUs, \textit{trainer worker groups}, which colocate the reference and actor models, pull data from the global buffer to compute weight updates, as consumers. This decoupling theoretically allows generation and training to proceed concurrently, so that the rollout phase of long-tail tasks can take place simultaneously with the training phase of shorter tasks.

While effective for single-turn tasks, current asynchronous architectures exhibit severe systemic inefficiencies when processing multi-turn agentic workloads characterized by high concurrency and variable context lengths. We identify three critical challenges:

\begin{icompact}
    \item \textbf{KV cache preemption during intermittent interactions.} Standard inference engines rely on request-level continuous batching, assuming each request is self-contained. However, multi-turn tasks proceed as a sequence of dependent requests separated by idle periods while waiting for external feedback. During these pauses,
    % When a multi-turn task pauses to wait for network-bound environment feedback, 
    the engine greedily evicts its KV cache (the memory footprint of historical tokens) to serve 
    requests from other concurrent tasks. Upon resumption, the system suffers a catastrophic cache miss as Figure~\ref{fig:cache_hitrate} shows, forcing redundant and computationally expensive recomputation (prefill) for an increasingly long historical context.
    \item \textbf{Stalls and thrashing in reference inference.} Existing systems share GPU resources between actor and reference workers within the trainer, enforcing strict stage dependencies. To avoid memory exhaustion, they either wait for large global batches to arrive before initiating updates, or frequently alternate between reference and actor models (which share the same architecture but have different weights) on the same GPUs for each micro-batch. The former creates massive computation bubbles that leave GPUs idle for up to 81\% of the step time, while the latter introduces severe model-swapping thrashing overhead.
    \item \textbf{Rigidity of resource allocation under shifting bottlenecks.} Agentic RL operates as a producer-consumer pipeline. Because task complexity varies drastically, the system bottleneck dynamically shifts. A batch of complex tasks stalls the trainer as it waits for slow rollout data, while simple tasks create a large ready backlog at the beginning of a training step and then keep feeding new micro-batches rapidly. Relying on static resource partitioning between rollout and trainer worker groups, existing systems cannot react to either the initial ready backlog or the steady ready interval, structurally capping global throughput during these macroscopic workload shifts, as shown in Figure~\ref{fig:resource_util}.
\end{icompact}

These intertwined challenges expose architectural flaws of existing systems, raising a critical research question: \textit{How can we build a unified, elastic training architecture that maximizes RL goodput, as measured by training throughput, under multi-turn agentic execution?} The key is not to optimize rollout, training, and resource allocation as isolated modules. Multi-turn RL needs a shared readiness view: Rollout must preserve the task state that will become useful soon, Trainer must choose a ref-actor execution mode that matches how ready data arrives, and the cluster scheduler must move GPUs toward the side that removes the current goodput loss. Answering this requires co-designing the multi-turn memory management, distributed execution pipeline, and dynamic cluster resource allocation.

To this end, we identify two architectural blind spots in existing frameworks. First, systems such as AReaL and StreamRL expose asynchrony, but still treat the rollout-trainer split as a mostly fixed operating point; they cannot use the current ready backlog and arrival pace to jointly choose the ref-actor execution strategy and the GPU split. Second, standard inference engines allocate memory at the \textit{request} level. Memory management must be elevated to the \textit{task} level to gain global visibility over the multi-turn lifecycle, actively coordinating context pinning across distributed rollout ranks. Together, these changes make individual data parallel~(DP) ranks schedulable resources rather than static members of a worker group.

Motivated by these insights, we present \sys, an elastic and asynchronous distributed system tailored for multi-turn agentic RL goodput. \sys solves these challenges through three co-designed components:
\begin{icompact}
    \item \textbf{Continuous Task Batching (CTB)} replaces request-level continuous batching on each rollout rank with task-level admission, preemption, and resumption. CTB tracks per-rank token footprints in real time and applies a semantic-aware priority hierarchy---prioritizing evaluation boundaries, near-complete GRPO groups, active trajectories, and long contexts. This keeps useful rollout state resident until it can feed the global buffer, providing the stable producer side that \rap and ERS rely on.
    %Concurrently, we address the severe KV cache preemption on the rollout side. Because standard continuous batching blindly evicts caches during intermittent environment feedback, CTB elevates memory management to the task level. By monitoring the real-time token footprint across individual rollout ranks, CTB selectively pauses and resumes tasks, ensuring that active multimodal KV caches remain securely pinned until the multi-turn episode naturally concludes.
    \item \textbf{Resource-Aware Ref-Actor Pipelining (\boldrap)} resolves the stall-thrashing dilemma through two complementary execution modes selected by two data-readiness signals: \textit{Ready-at-Start} (RAS), the amount of micro-batch work already present when a training step begins, and \textit{Time Per Ready Micro-batch} (TPRM), the interval at which additional micro-batches become available. In decoupled mode, Ref and Actor run on separate GPUs in a streaming micro-batch pipeline; a restructured computation graph defers loss computation into the backward pass, enabling Ref forward, Actor forward, and Actor backward to overlap with zero swap overhead. In colocated mode, \rap uses ready-batch aggregation and zero-copy shared-memory transfer to amortize unavoidable swapping across ready bursts. Together, \rap turns readiness into goodput under both startup backlog pressure and steady per-micro-batch arrival pressure.
    % To overcome the stalls and thrashing in reference inference, we introduce \rap with two execution modes. When a global batch contains few large micro-batches, \rap operates in a \textit{colocated mode}. It utilizes zero-copy transmission to eliminate data serialization and deserialization costs, making the inevitable model-swapping overhead manageable. Conversely, when a global batch comprises many micro-batches, the cumulative overhead of frequent model loading and offloading becomes prohibitive. In this scenario, \rap switches to a \textit{decoupled pipeline mode}, deploying reference and actor models on separate GPUs to completely eliminate model-swapping thrashing and overlap reference inference with policy updates.
    \item \textbf{Elastic Resource Scaling (ERS)} exploits \rap's dual-mode structure to migrate individual ranks between rollout and training on the fly. ERS reads RAS and TPRM from the global buffer: a small RAS with slow TPRM indicates rollout starvation and triggers colocated \rap plus more Rollout ranks, whereas a large RAS or fast TPRM indicates trainer-side pressure and triggers decoupled \rap plus more Trainer capacity. Crucially, ERS piggybacks weight migration onto the natural cache-flush boundary that on-policy RL already mandates at every sync, achieving zero-overhead elasticity: no pipeline suspension, no NCCL group rebuild, no wasted KV cache.
    % The dual-mode design of \rap structurally allows trainer ranks to shrink (colocated) and expand (decoupled) safely, creating the prerequisite for ERS. To combat resource rigidity, ERS dynamically rebalances compute power between generation and training on the fly. For the first time, we utilize the \textit{rollout data waiting time in the global buffer}, combined with partial-step evaluations, as a sensitive scheduling metric. Guided by this metric, ERS seamlessly commands individual ranks to switch \rap modes and migrate GPUs, absorbing macroscopic workload shifts and unlocking previously trapped cluster capacity.
\end{icompact}

% We implement \sys with a highly modular architecture, ensuring seamless compatibility with different types of multi-modal or text-only tasks, and easy integration with mainstream serving and training backends. We deployed the system on an 4-node cluster with 32 NVIDIA H100 GPUs. We train text-only and multi-modal large models of 4B--27B parameters on practical tasks. Our results demonstrate significant performance gains over existing asynchronous frameworks. \sys increases first-round training throughput by over 38\%. Specifically, runtime elasticity via ERS yields an additional 10\% improvement by rescuing idle compute nodes. Furthermore, CTB effectively mitigates context preemption, boosting the multi-turn prefix cache hit rate by 1.8$\times$ and throughput by 13\%.

We implement \sys with a highly modular architecture, ensuring seamless compatibility with diverse multi-modal and text-only tasks, and easy integration with mainstream serving and training backends. We deploy the system on a 4-node cluster with 32 NVIDIA H100 GPUs and evaluate models with different architectures. Our evaluations demonstrate large performance gains over existing RL frameworks. End-to-end, \sys raises RL training goodput, measured by training throughput, by 1.8--7.0$\times$ on text-only tasks (reducing time-to-convergence by 51.1\%) and improves multi-modal goodput by over 33\%, reducing training time by 62.2\% with similar performance. Component studies close the loop from motivation to design: CTB improves generation goodput by preserving task state, \rap reduces per-step training time by up to 44.3\% under different readiness patterns, and ERS cuts total idle waiting time by 87.4\% by scheduling \rap modes and GPU ranks from the same readiness signals.

\section{Background and Motivation}
\label{sec:bg}
We first summarize the agentic RL workflow and then motivate the three system bottlenecks that \sys targets.

\subsection{Agentic Reinforcement Learning}
Agentic RL workloads in this paper follow Group Relative Policy Optimization (GRPO)~\cite{grpo}, whose execution alternates between trajectory generation and model update.

\noindent \textbf{The Agentic RL Pipeline.}
Unlike traditional single-turn chat optimization, agentic RL requires models to interact with environments iteratively. As depicted in Figure~\ref{fig:concept_arch}(b), a standard agentic GRPO pipeline consists of four sequential stages to train a new model $\mathcal{A}$ from the original model $\mathcal{O}$:
\begin{icompact}
    \item \textbf{Rollout Generation:} Rollout\footnote{For clarity and brevity, \textit{Rollout}, \textit{Ref}, and \textit{Actor} denote the worker group of rollout worker, reference worker, and actor worker, respectively, and \textit{Env} denotes environment, throughout the remainder of this paper.} hosts $\mathcal{A}$ to generate actions. Crucially, this involves \textit{multi-turn} interactions: the model generates an action, pauses to wait for environment execution, receives the observation, and generates the next action. For a given task, a group of $G$ distinct multi-turn trajectories is sampled. Each trajectory may comprise different turns and lengths of messages.
    \item \textbf{Reward Calculation:} Once the trajectories in a group are completed, the environment evaluates them. GRPO determines the advantage $A_i$ by standardizing the rewards $r_i$ strictly based on the performance of other trajectories within the same group: $A_i = (r_i - \text{mean}(\mathbf{r})) / \text{std}(\mathbf{r})$.
    \item \textbf{Reference Calculation:} Ref computes the base log-probabilities on $\mathcal{O}$ of generated trajectories. This provides a Kullback-Leibler (KL) divergence penalty to prevent the policy from over-optimizing the reward.
    \item \textbf{Actor Update:} Actor executes the forward and backward passes on $\mathcal{A}$ to update its weights (\textit{i.e.}, update the policy). Once updated, the latest weights are synchronized back to Rollout for the next iteration.
\end{icompact}

The \textit{Trainer} comprises Ref, Actor and the reward calculation, and consumes the trajectories from Rollout.

\noindent \textbf{Asynchronous RL.}
Traditional synchronous architectures enforce a strict alternation between the rollout and training stages. This coupling causes severe GPU idling, as the training must wait for the long-tail task to complete~\cite{openrlhf, rlhfuse}. To break this resource coupling, modern systems typically adopt an asynchronous execution paradigm~\cite{streamrl,areal}. As depicted in Figure~\ref{fig:concept_arch}(c), they decouple generation and training into physically independent workers. This allows Rollout to continuously generate new groups of responses while Trainer consumes historical data in the background, significantly boosting hardware throughput. However, this asynchrony introduces \textit{policy staleness}, as Rollout generates data using older model weights. Since algorithms like GRPO are fundamentally on-policy, excessive off-policy data severely degrade training stability. Consequently, asynchronous systems must strictly bound this staleness gap, requiring timely weight synchronization to prevent performance degradation.

\subsection{Key Challenges in Asynchronous Agentic RL}
\label{sec:system_challenges}

\begin{figure}[tb]
\centering
\includegraphics[width=0.95\linewidth]{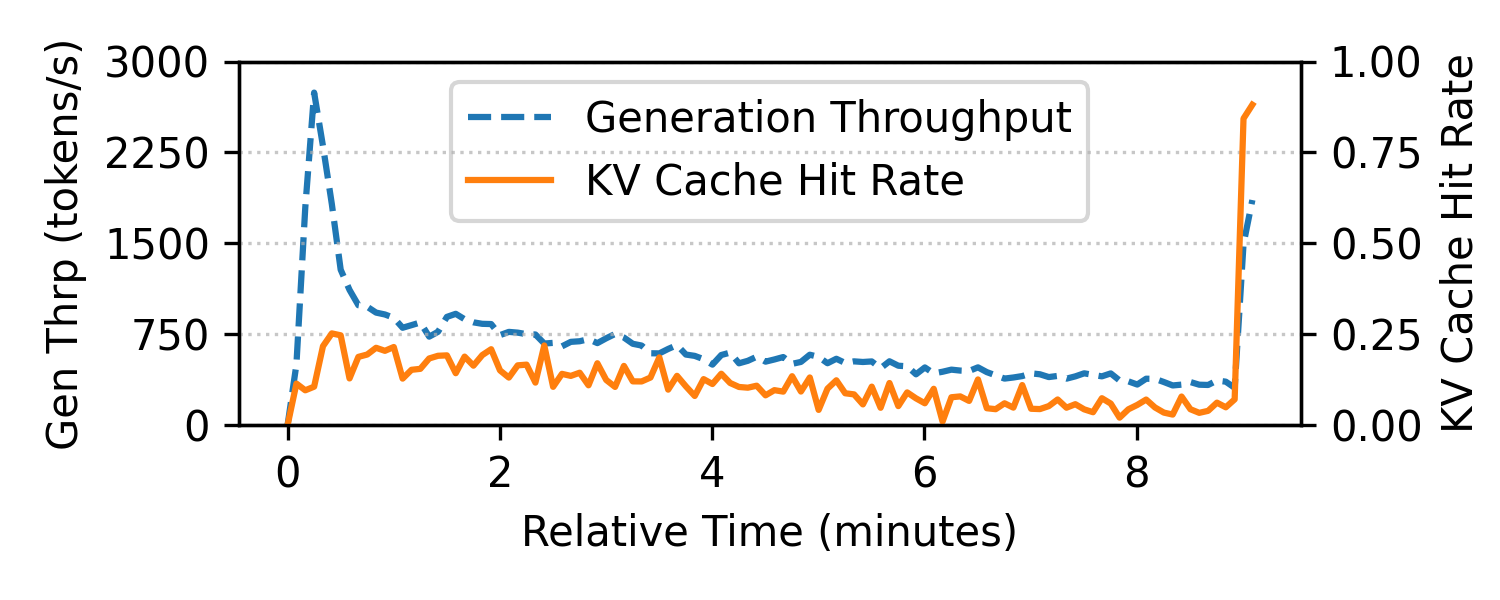}
\caption{Multi-turn agentic tasks experience severe cache misses in one WebShop training step. The data is collected on one NVIDIA H100 GPU running Qwen2.5-7B~\cite{qwen25}.}
\label{fig:cache_hitrate}
\end{figure}

\begin{figure}[tb]
    \centering
    \includegraphics[width=\linewidth]{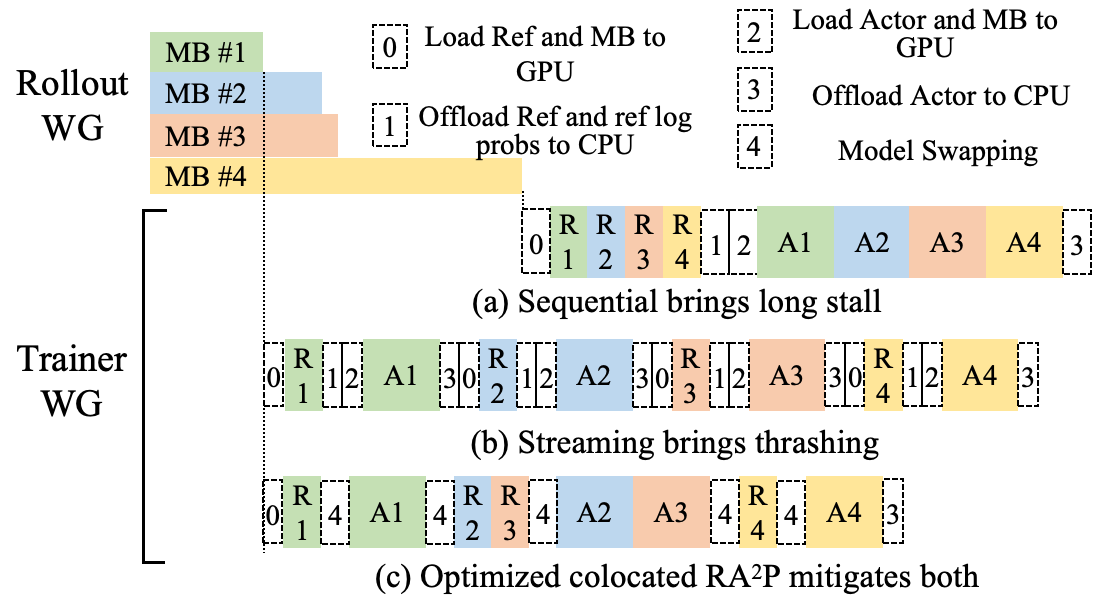}
    \caption{Trainer experiences significant stalls and thrashing due to long-tail tasks. Even without CPU offloading, model alternation still incurs non-negligible GPU context-switching overhead, as shown in~\cref{sec:eval:extend}.}
    \label{fig:pipeline_problems}
\end{figure}

\begin{figure}[tb]
    \includegraphics[width=1.0\linewidth]{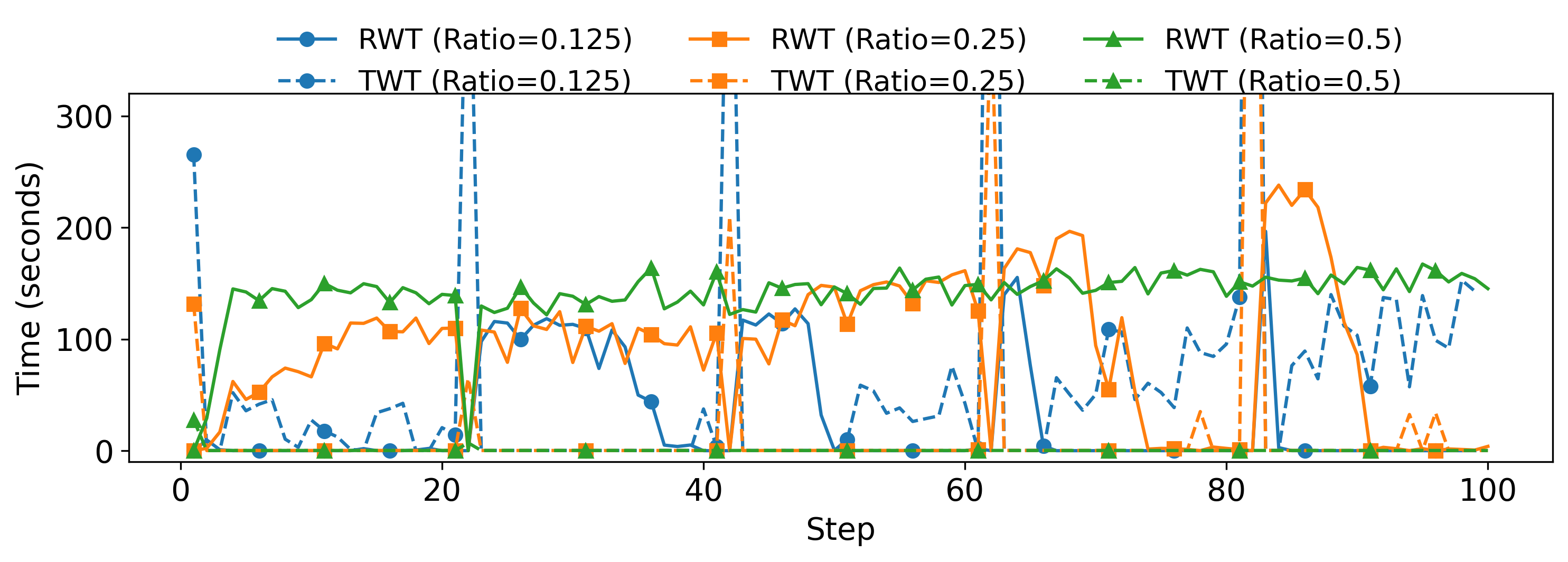}
    \caption{Rollout Wait Time~(RWT) and Training Wait Time~(TWT) during the first 100 steps under varying fraction of GPUs assigned to Rollout. These waits expose the two readiness signals that drive \rap and ERS. The experiment is carried out on one node with 8 NVIDIA H100 GPUs, running the WebShop~\cite{webshop} agents.}
    \label{fig:resource_util}
\end{figure}

Although asynchronous RL frameworks overlap generation and training, their execution traces under multi-turn agentic workloads reveal three ways the system loses RL goodput.

\vspace{1ex}
\noindent\textbf{C1: Catastrophic KV Cache Preemption in Rollouts.} \textit{Observation:} Shifting our focus to the Rollout phase, we monitored the KV cache hit rate and generation throughput during a single training step. Figure~\ref{fig:cache_hitrate} exhibits a typical three-phase fluctuation rather than stable cache utilization. At the beginning of the step, the hit rate increases as a large influx of new short sequences is initially cached. However, the hit rate subsequently plummets and remains at only $\sim12\%$, dragging the throughput down to $\sim650$ tokens/s$\cdot$instance. A spike in the hit rate occurs at the very end of the training step.

\noindent\textit{Root Cause:} The fundamental flaw lies in the mismatch between request-level scheduling and multi-turn semantics. While recent scheduling optimizations target single-turn reasoning tasks, they fall short in multi-turn agentic workloads. For instance, Seer's~\cite{seer} chunk-level scheduling yields minimal benefits because agentic outputs per request are typically short, and StreamRL's~\cite{streamrl} length-probing mechanism fails to capture the intermittent nature of environment interactions~\cite{agserve, continuum}. When waiting for environment feedback, the underlying inference engine, driven by standard continuous batching to maximize concurrency, schedules new requests, thereby preempting the tasks' KV caches. The engine suffers massive cache misses as the observation returns, and must recompute the rapidly growing historical context. The late recovery in hit rate is merely an artifact of dropping concurrency: as the global batch nears completion, no new tasks are admitted, allowing the few straggling tasks to safely retain their cache without preemption.

\vspace{1ex}
\noindent\textbf{C2: Stalls and Thrashing in Training.} \textit{Observation:} We examined the Trainer's execution trace, and observed a dilemma between computation stalls and I/O thrashing. In Figure~\ref{fig:pipeline_problems}(a), standard execution forces the Trainer to stall significantly while waiting for a full global batch to complete, a delay severely exacerbated by long-tail tasks. Conversely, if we attempt to mitigate this idle time by streaming micro-batches, the system falls into another trap: massive I/O thrashing that degrades training GPU goodput, as illustrated in Figure~\ref{fig:pipeline_problems}(b).

\noindent\textit{Root Cause:} This dilemma is structurally rooted in the memory-intensive nature of on-policy algorithms like GRPO. The Trainer must execute reference calculations and actor updates sequentially using two distinct models. Under standard execution, the system waits for all rollout tasks in a global batch to finish, making the Trainer's idle time bottlenecked by the slowest long-tail samples. To avoid these macroscopic stalls, the system could adopt a streaming mode to process micro-batches as soon as they arrive. However, existing systems (\eg, VeRL~\cite{verl} and AsyncFlow~\cite{asyncflow}) cannot run both models on the same GPUs simultaneously due to the runtime memory consumed by training activations. Consequently, streaming micro-batches forces the system to frequently load and offload model weights between the CPU and GPU over PCIe, replacing the stall with a model-swapping bottleneck.

\vspace{1ex}
\noindent\textbf{C3: Shifting Bottlenecks Invalidating Static Allocation.}

\noindent\textit{Observation:} To investigate whether global resource tuning could improve agentic RL goodput, we tracked the Rollout Wait Time (RWT) and Training Wait Time (TWT) across 100 steps under various static Rollout GPU ratios. Since \sys measures goodput through training throughput, these waits matter because they directly consume time without advancing useful training. Fundamentally, RWT measures the idle duration Rollout workers stall upon reaching off-policy staleness limits and waiting for new weights, whereas TWT quantifies the Trainer's starvation period awaiting sufficient trajectory accumulation. These two waits correspond to the Trainer's data-readiness state: high TWT means the step begins with too few ready micro-batches and receives later ones slowly, while high RWT means Rollout has produced a backlog that the Trainer cannot consume quickly enough. As illustrated in Figure~\ref{fig:resource_util}, the system bottleneck oscillates wildly. A lower rollout ratio (0.125) consistently starves the Trainer (high TWT), while a higher ratio (0.5) generates data too fast, causing Rollout to stall (high RWT). Most crucially, even within a single optimal fixed ratio (\eg, 0.25), RWT and TWT frequently cross paths.

\noindent\textit{Root Cause:} This invalidation of static partitioning stems from the extreme variance in agentic trajectory execution times, further compounded by the unpredictable load of intermediate evaluation tasks. Unlike single-turn chat, multi-turn environments cause the computational bottleneck to dynamically flip between generation and training, so a static GPU quota cannot preserve goodput across steps. It cannot react to either the micro-batch backlog already ready at a training boundary or the per-micro-batch ready interval after the step starts. These two signals determine whether the trainer should spend GPUs on a decoupled Ref-Actor pipeline or release them to Rollout through colocated execution. Although StreamRL~\cite{streamrl} attempts to address this by dynamically increasing the size of Rollout, it lacks real-time reaction due to high initialization overheads. Furthermore, this scaling approach is incompatible with the fixed-resource environments prevalent in large-scale production clusters, such as those operated by Alibaba and Microsoft~\cite{antman, pollux, gao2024empirical}. As a result, the cluster is left severely underutilized throughout the training lifecycle, with goodput lost to both data starvation and off-policy throttling.

\vspace{1ex}
\noindent\textbf{Summary.} The paradigm shift toward multi-turn agentic RL exposes fundamental architectural mismatches. Request-level scheduling leads to severe cache preemption (\textbf{C1}), rigid stage dependencies create computation bubbles (\textbf{C2}), and static partitioning fails to adapt to shifting bottlenecks (\textbf{C3}), all of which reduce training throughput and therefore goodput. Addressing these inefficiencies necessitates a holistic system redesign, spanning from computation graph execution to physical resource allocation and memory management.

\section{\sys Overview}\label{sec:overview}

To address these challenges, we propose \sys, an elastic asynchronous RL system tailored for multi-turn agentic workloads. \sys coordinates three decisions that existing systems handle separately: which tasks stay resident in Rollout, how ready trajectories are consumed by the Trainer, and how GPUs move between the two sides.

\subsection{Design Rationale}
\label{sec:design_rationale}

The design follows the dataflow of agentic RL. CTB keeps Rollout productive without losing task context; \rap consumes ready micro-batches without waiting for a full global batch or thrashing between Ref and Actor; ERS adjusts resources so the selected \rap strategy matches the current readiness pattern. The three components are intentionally coupled by one feedback loop: CTB controls when trajectories become ready, \rap exposes how expensive it is to consume the current ready stream, and ERS moves ranks to reduce the dominant wait revealed by that stream.

% \noindent\textbf{Ref-Actor Pipelining ($\mathrm{RA}^2\mathrm{P}$).} To eliminate the pipeline bubbles caused by synchronous reference evaluation (\textbf{C1}), \sys restructures the computation graph on the training side. Instead of waiting for the completion of a global batch, $\mathrm{RA}^2\mathrm{P}$ separates Ref and Actor onto different GPUs, and groups one Ref rank with multiple Actor ranks to form a micro-batch pipeline. As Rollout generates trajectories, data is streamed directly to the $\mathrm{RA}^2\mathrm{P}$ group, allowing the Ref to compute the KL divergence penalty concurrently with the Actor's forward and backward passes. This streaming architecture completely bypasses the global synchronization barrier.

\noindent\textbf{Continuous Task Batching (CTB).} To resolve KV cache preemption in multi-turn interactions (\textbf{C1}), \sys elevates Rollout scheduling from requests to tasks. CTB tracks each task's token footprint, admits tasks only when a rank has enough KV cache headroom, pauses tasks according to RL-aware priorities, and resumes them with worker affinity. This preserves useful context while maintaining high concurrency.

\noindent\textbf{Resource-Aware Ref-Actor Pipelining (\boldrap).} To address Trainer stalls and model alternation thrashing (\textbf{C2}), \sys streams generated trajectories directly to the Trainer. \rap then chooses between two Ref-Actor execution strategies. When ready micro-batches are abundant or arrive quickly, \rap decouples Ref and Actor onto different GPUs to overlap reference computation with policy updates. When ready micro-batches are scarce, \rap colocates Ref and Actor, uses ready-batch aggregation and zero-copy transfer to amortize model alternation, and leaves more GPUs available for Rollout.

\noindent\textbf{Elastic Resource Scaling (ERS).} To absorb shifting bottlenecks (\textbf{C3}), \sys abandons static resource partitioning. ERS acts as the runtime scheduler for the two \rap strategies: it provisions more Rollout ranks and selects colocated \rap when the Trainer is data-starved to reduce TWT, and it switches to decoupled \rap with additional Trainer capacity when ready data accumulates to reduce RWT. This keeps Ref-Actor execution aligned with the current producer-consumer imbalance.

\subsection{System Architecture and Workflow}
\label{sec:system_workflow}

\begin{figure}[t]
\centering
% Placeholder for the detailed system architecture and data flow diagram
\includegraphics[width=\linewidth]{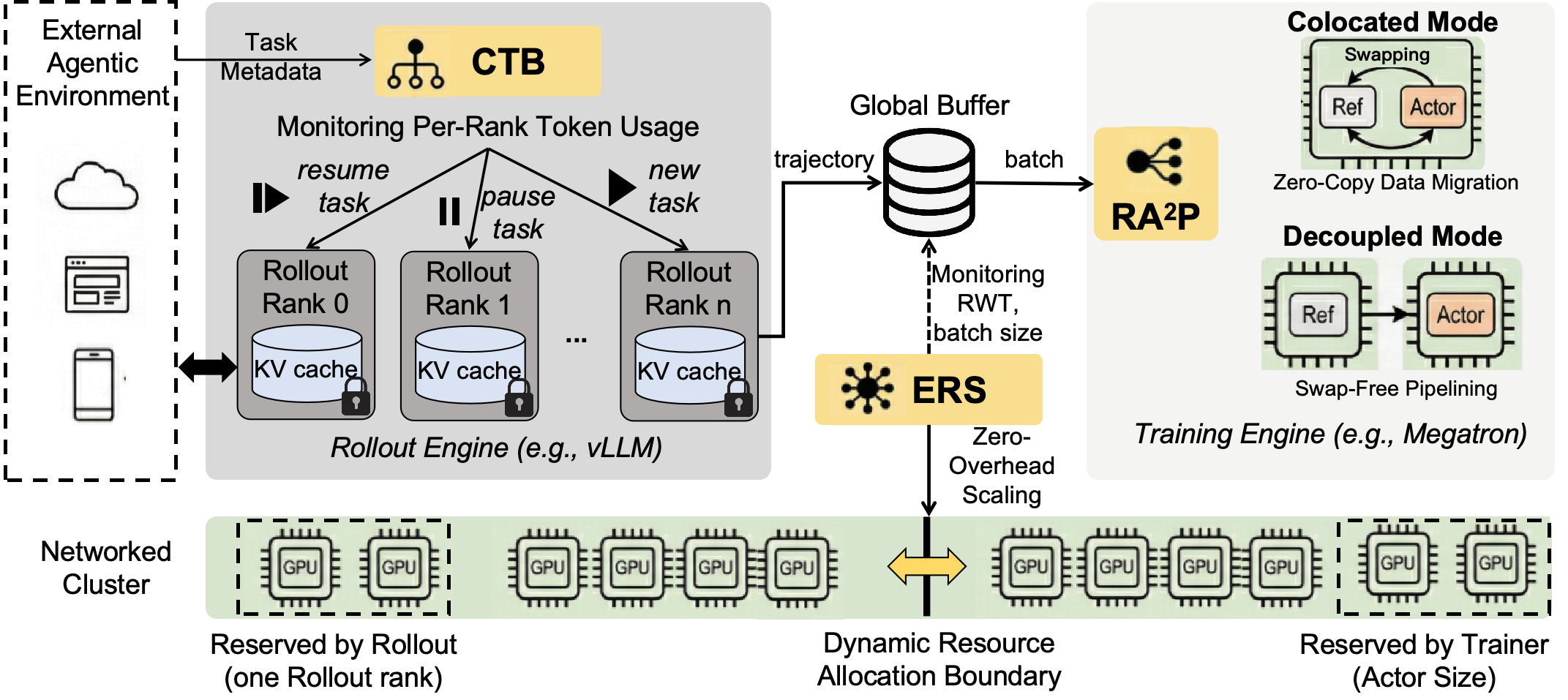}
\caption{The overall architecture and workflow of \sys. Environments and reward computations are disaggregated to an external CPU cluster. The system features a continuous data flow orchestrated by CTB on Rollout ranks, \rap on Trainer nodes, and an overarching ERS scheduler.}
\label{fig:sys_architecture}
\end{figure}

Inspired by AgentRL~\cite{agentrl}, \sys adopts a disaggregated architecture where environment maintenance and reward computation are offloaded to an external CPU cluster. As illustrated in Figure~\ref{fig:sys_architecture}, this isolates GPU instances purely for rollout and training. We trace the continuous lifecycle of a task and its trajectory to illustrate \sys's workflow.

A task's lifecycle begins in the background, where CTB manages the admission of new agentic tasks based on Actor KV cache availability and strictly bounds off-policy staleness. Once dispatched to a Rollout rank, the Actor and Env engage in multi-turn interactions. During this highly concurrent multi-task generation, CTB monitors the real-time token footprint, pausing or resuming tasks to achieve higher throughput and prevent memory exhaustion.

Upon completing all interaction turns, the Env calculates group-relative rewards and pushes the group to a global rollout buffer. The Trainer pulls ready trajectories from the buffer and uses \rap to update the policy with low or no model-swapping overhead. Once the training step completes, Trainer synchronizes the new weights to Rollout.

The overarching ERS scheduler monitors the utilization and readiness state of the rollout buffer. If trajectory generation outpaces training, ERS dynamically commands a subset of Rollout ranks to safely offload their Actor weights and reconfigures these freed nodes into the trainer group; if training starves, it moves trainer-side ranks back to Rollout and lets \rap run in the colocated mode.

\sys also considers evaluation tasks. For the evaluation of the $i$-th step's model, CTB organizes them concurrently with training data rollout in the $(i+1)$-th step. CTB aggregates the metrics just before the weight synchronization of the $(i+1)$-th step. This ensures metrics are reported accurately without interrupting the continuous trajectory flow.

\section{Continuous Task Batching}
\label{sec:ctb}

CTB makes Rollout scheduling task-aware. Instead of treating every environment turn as an independent request, CTB tracks each task's growing context, admits new tasks only when a Rollout rank has enough KV cache headroom, pauses tasks before memory is exhausted, and resumes paused tasks with cache locality whenever possible.

\subsection{Token-Aware Admission Control}
In highly concurrent agentic RL, dispatching tasks by raw request counts creates severe imbalance because tasks consume very different amounts of KV cache. CTB instead treats each Data Parallel (DP) rank as a token budget pool and admits tasks according to measured context size.

To estimate a task's initial footprint, CTB sends a lightweight probing task for each trajectory group and records the token usage of its first turn, following the idea of Seer~\cite{seer}. During dispatch, CTB checks the current token utilization of each rank and admits a task only if the remaining headroom can accommodate the probed footprint. After admission, CTB updates the task's actual token usage at the start of every interaction turn, so later scheduling decisions reflect the growing context rather than the initial estimate.

CTB also enforces three admission constraints:
\begin{icompact}
    \item \textbf{Environment concurrency:} Active tasks cannot exceed the parallel capacity of the external environments.
    \item \textbf{Policy staleness:} CTB throttles dispatch when Rollout is likely to generate more data than Trainer can consume within the staleness bound.
    \item \textbf{Task diversity:} CTB penalizes assigning too many identical task types to the same batch, spreading environment-side stragglers across ranks.
\end{icompact}

\begin{algorithm}[tb]
\caption{Semantic-Aware CTB Scheduling}
\label{alg:ctb}
\textbf{Input:} Active $\mathcal{T}_{act}$, Paused $\mathcal{T}_{pause}$, New $\mathcal{T}_{new}$, Rank budgets $\mathcal{B}$
\begin{algorithmic}[1]
\small
\WHILE{$\mathcal{T}_{act}\cup\mathcal{T}_{pause}\cup\mathcal{T}_{new}\neq \emptyset$}
    \STATE Update token footprint for $t \in \mathcal{T}_{act}$ via turn probing
    \FOR{each rank $r$}
        \WHILE{TokenUsage($r$) $> \mathcal{B}_r$}
            \STATE $t_{evict} \leftarrow \text{argmin}_{t \in \mathcal{T}_{act}^r} \text{Priority}(t)$ \COMMENT{Tiers 1-4}
            \STATE Preempt $t_{evict}$ from $\mathcal{T}_{act}^r$ to $\mathcal{T}_{pause}$
        \ENDWHILE
        \WHILE{TokenUsage($r$) $< \mathcal{B}_r$}
            \IF{$\mathcal{T}_{pause} \neq \emptyset$ \AND fits in $\mathcal{B}_r$}
                \STATE $t_{res} \leftarrow \text{argmax}_{t \in \mathcal{T}_{pause}} \text{Priority}(t)$
                \STATE Resume $t_{res}$ on $r$ (enforcing group affinity)
            \ELSIF{$\mathcal{T}_{new} \neq \emptyset$ \AND satisfies constraints}
                \STATE Admit $t_{new}$ to $r$ based on probed footprint
            \ELSE
                \STATE \textbf{break}
            \ENDIF
        \ENDWHILE
    \ENDFOR
\ENDWHILE
\end{algorithmic}
\end{algorithm}

\subsection{Semantic-Aware Pausing and Resuming}
As tasks progress through multiple turns, their contexts may outgrow a rank's token budget. CTB handles this by pausing selected tasks before KV cache pressure causes uncontrolled eviction. Algorithm~\ref{alg:ctb} summarizes the loop: update active footprints, pause low-priority tasks when a rank exceeds its budget, and fill recovered headroom by resuming paused tasks before admitting new ones.

The priority score combines RL progress with system cost:
\begin{equation*}
    P(t) = \omega_1 \cdot \mathbb{I}_\mathrm{eval} + \omega_2 \cdot G_{\mathrm{completion}} + \omega_3 \cdot \mathbb{I}_\mathrm{active} + \omega_4 \cdot L_\mathrm{context}
\end{equation*}

Here, $\mathbb{I}_{\mathrm{eval}}$ marks evaluation tasks, $G_{\mathrm{completion}}$ is the completion ratio of the task's GRPO group (\eg, $N_{\mathrm{finished}}/N_{\mathrm{total}}$), $\mathbb{I}_{\mathrm{active}}$ marks tasks currently executing, and $L_{\mathrm{context}}$ is the current token length. Table~\ref{tab:preemption_taxonomy} summarizes why each term matters.

\begin{table}[tb]
\centering
\footnotesize
\caption{Taxonomy of CTB's semantic-aware preemption hierarchy.}
\label{tab:preemption_taxonomy}
\setlength{\tabcolsep}{3pt} % Reduces horizontal padding slightly for tight columns
\begin{tabularx}{\columnwidth}{@{} l X l @{}}
\toprule
\textbf{Metric} & \textbf{Rationale} & \textbf{Priority} \\
\midrule
Task Type & Evaluation tasks require anchored weights; stalling them blocks global weight synchronization. & \textbf{Highest} \\
\addlinespace
Group Comp. & GRPO requires full groups to calculate advantages; fragmented groups stall gradient updates. & \textbf{High} \\
\addlinespace
Exec. State & Preempting actively running tasks destroys in-flight compute cycles and prior investments. & \textbf{Med.} \\
\addlinespace
Context Len. & Evicting massive historical contexts causes catastrophic redundant prefill upon resumption. & \textbf{Low} \\
\bottomrule
\end{tabularx}
\vspace{2ex}
\end{table}

When capacity is restored, CTB resumes paused tasks before admitting new tasks. It also enforces \textit{worker affinity} based on group IDs, placing resumed tasks on ranks that are most likely to retain their shared prefixes. This preserves useful KV cache state and avoids unnecessary prefix recomputation.

\section{Resource-Aware Ref-Actor Collaboration}
\label{sec:rap}

The Trainer must process rollout data as soon as it becomes useful, but agentic workloads make data readiness highly uneven. At some steps, many micro-batches are already waiting in the global buffer; at others, the Trainer starts almost empty and receives new micro-batches slowly.

\sys uses RAS (startup ready backlog) and TPRM (later ready interval) in two layers. First, \rap provides two Ref-Actor execution strategies: a decoupled streaming strategy for high RAS or short TPRM, and a colocated aggregation strategy for low RAS and long TPRM. Second, ERS schedules between these two \rap strategies by moving ranks between Rollout and Trainer. This section first explains the two \rap strategies and then shows how ERS selects and provisions them.

\subsection{\texorpdfstring{\boldrap}{RA2P} Execution Strategies}

\begin{figure}[tb]
    \centering
    \includegraphics[width=0.85
    \linewidth]{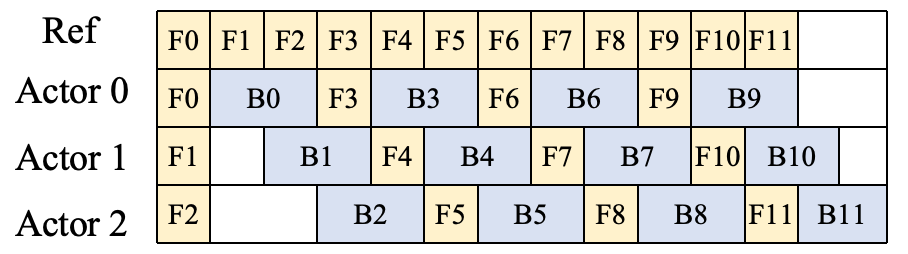}
    \caption{The \rap pipeline for decoupled streaming mode. The loss computation for Actor ranks is moved from the forward pass into the backward pass.}
    \label{fig:rap_pipeline}
\end{figure}

\vspace{0.45ex}
\noindent\textbf{Decoupled Streaming Mode.} 
When RAS is large or TPRM is short, the Trainer has enough ready work to keep dedicated Ref and Actor resources busy. \rap therefore avoids model-swapping overhead by allocating the Reference (Ref) and Actor models onto distinct GPUs. Since Ref only performs forward passes, \rap groups one Ref rank with multiple Actor ranks to form a streaming micro-batch pipeline.

To maximize concurrency and avoid blocking the Actor, we restructure the computation graph. In standard implementations, the loss computation is included in the Actor's forward pass, requiring Actor to wait for the Ref to output the reference log probabilities (\texttt{ref\_log\_probs}) before the forward pass. \rap breaks this dependency by deferring the Actor's loss computation (including the KL divergence penalty) to its backward pass. As shown in Figure~\ref{fig:rap_pipeline}, the Ref model sequentially computes \texttt{ref\_log\_probs} ($F_0, F_1, F_2$), while the Actor performs its forward passes ($F_0, F_3, F_6$) independently. The Actor only consumes the \texttt{ref\_log\_probs} when initiating the backward passes ($B_0, B_3, B_6$). This realignment ensures overlap between reference log prob computation and policy updates with much lower synchronization bubbles.

\vspace{0.45ex}
\noindent\textbf{Optimized Colocated Mode.}
When RAS is small and TPRM is long, a dedicated Ref rank would spend much of the step waiting for data. \rap then colocates the Ref and Actor models on the same GPUs, releasing the extra GPUs to Rollout where they can shorten TPRM for later steps. To keep colocation efficient, \rap introduces two optimizations, as depicted in Figure~\ref{fig:pipeline_problems}(c).

First, we implement \textit{ready-batch aggregation}. Instead of alternating between models for every single micro-batch (\eg, $R_1 \rightarrow A_1 \rightarrow R_2 \rightarrow A_2$), \rap monitors the global rollout buffer. If multiple micro-batches are ready, it aggregates their execution. The system loads the Ref model once to process all ready batches ($R_{2,3}$), and then swaps to the Actor to perform updates ($A_{2,3}$). This amortizes the I/O cost of model alternation. 

Second, \rap leverages \textit{zero-copy transmission}. Because the Ref and Actor reside on the same physical node, the input sequences and the generated \texttt{ref\_log\_probs} are passed directly via GPU shared memory, bypassing network serialization and redundant memory allocations.

\vspace{0.45ex}
\noindent\textbf{Analytical Mode Selection.}
The two strategies target different bottlenecks. Decoupled mode pays a pipeline-fill cost but removes model swaps, so it is best when high RAS amortizes the fill cost or short TPRM keeps the pipeline continuously fed. Colocated mode keeps fewer Trainer GPUs active and uses ready-batch aggregation to amortize occasional swaps, so it is best when low RAS and long TPRM would otherwise leave a decoupled Ref rank idle. \rap compares the decoupled latency cost ($C_{dec}$) and colocated latency cost ($C_{col}$) under these two signals; Appendix~\ref{sec:appendix_rap_cost} details the cost model.

\vspace{0.5ex}
\noindent\textbf{Micro-Batch Dispatching.}
To maximize throughput and prevent stragglers, \rap jointly optimizes \textit{how} a micro-batch is formed and allocated across DP ranks.

\rap employs the same micro-batching strategy for both execution modes. Early micro-batches are accumulated from the rollout buffer until they contain sufficient tokens to fully saturate the compute capacity of all DP ranks. Conversely, the last micro-batch of a global batch is deliberately kept as small as possible. This intentionally minimizes the duration of the final forward and backward passes, thereby reducing the pipeline flush bubble at the end of the training step.

\rap further applies distinct dispatching strategies to handle the heterogeneous micro-batch size:
\begin{icompact}
    \item \textit{Colocated Mode.} In this mode, GPU execution is purely sequential. The primary goal is to balance the total compute load across all training ranks. \rap utilizes a zig-zag (longest-processing-time-first) allocation strategy, sorting sequences by length and distributing them iteratively in a zig-zag pattern. This ensures that the sum of sequence lengths assigned to each rank remains nearly identical.
    \item \textit{Decoupled Mode.} In the decoupled mode, the overall completion time is bottlenecked by the execution time of the final micro-batch. Building upon our small-final-batch sizing strategy, \rap sorts and allocates the sequences such that the \textit{last} sequence fed into the pipeline is the absolute smallest. This minimizes the tail latency, allowing the last active rank to finish its backward pass at the earliest possible time.
\end{icompact}

\subsection{ERS Scheduling of \texorpdfstring{\boldrap}{RA2P} Strategies}
\label{sec:ers}
ERS closes the loop between \rap's mode choice and the producer-consumer imbalance. It reallocates GPUs on the fly so that Rollout can generate data fast enough and Trainer can consume ready micro-batches with the right \rap strategy.

\begin{figure}[tb]
    \centering
    \includegraphics[width=0.95\linewidth]{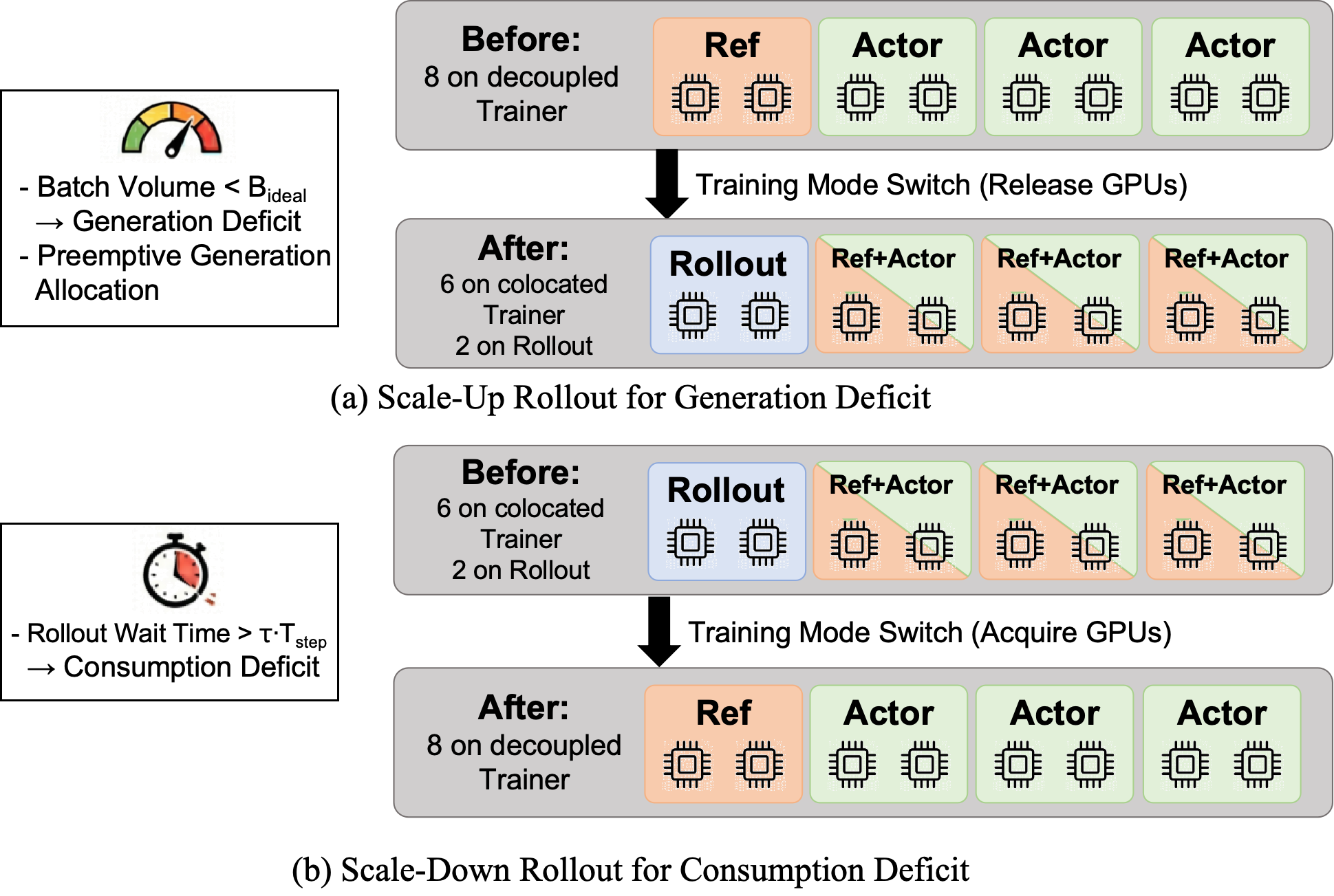}
    \caption{The ERS architecture. The ERS coordinator dynamically reallocates GPU resources between functional roles.}
    \label{fig:ers_arch}
\end{figure}

\vspace{0.45ex}
\noindent\textbf{Readiness-Aware Plan Generation.}
To accommodate volatile generation rates, \sys adapts the elastic batching strategy, where the Trainer consumes any accumulated batch size falling within an interval $[B_{min}, B_{ideal}]$. Here, $B_{min}$ is the minimum batch size required for an effective training step, while $B_{ideal}$ represents the full utilization of GPU HBM and computing resources. ERS generates scaling plans immediately after the Trainer finishes an Actor update, before the new weights are broadcast (\texttt{sync\_params}). 

Despite CTB and \rap, static resource allocation cannot perfectly align the generation and consumption throughputs. The mismatch appears as TWT when the Trainer starves for ready trajectories, and as RWT when Rollout is throttled by stale weights or an overfilled buffer. ERS monitors the global buffer to estimate the RAS and TPRM signals introduced above, and converts them into two complementary actions: scaling Rollout up to reduce TWT, and scaling Trainer up to reduce RWT.

Directly using instantaneous arrivals yields a noisy signal skewed by trajectory variance even within one step. To avoid reactive thrashing, the ERS coordinator (Figure~\ref{fig:ers_arch}) estimates these trends through two queue metrics: the generation and consumption deficits.

When Rollout resources are insufficient, generation slows. The next training step begins with a small RAS and observes a long TPRM, so the Trainer is starved both at startup and during the step, increasing TWT. ERS detects this via the \textit{generation deficit} metric (the global batch volume $< B_{ideal}$). It then selects colocated \rap and reallocates the GPUs previously used by decoupled Ref ranks to Rollout. This gives generation more capacity, directly reducing TWT, while the Trainer avoids wasting dedicated Ref GPUs on an empty stream.

Conversely, when Rollout is over-provisioned, ready data accumulates. The next training step starts with a large RAS, and continued fast generation shortens TPRM. If the Trainer cannot drain this data fast enough, Rollout eventually waits for buffer clearance and fresh weights, increasing RWT. ERS detects this \textit{consumption deficit} through the Head-of-Line (HoL) latency of the oldest micro-batch in the global buffer, together with the current ready volume. If the HoL latency exceeds a threshold ($\tau \times T_{step}$, where $T_{step}$ is the average step duration) or RAS exceeds the decoupled-mode break-even point, ERS selects decoupled \rap and reclaims a Rollout rank to provision a dedicated Ref rank. The Trainer can then drain the backlog without repeated model swaps, directly reducing RWT.

Beyond standard training, \sys handles boundary conditions via \textit{preemptive generation allocation}. During cold starts or evaluation phases, generation dominates. Rather than waiting for RAS to remain low and TPRM to become long, the coordinator preemptively maximizes the Rollout group size, allocating maximum hardware to bootstrap the buffer quickly.

\sys adopts \textit{adaptive data retention} to prevent starvation cycles following a Rollout scale-down. If a scaled-down Rollout cannot sustain the Trainer's consumption, the Trainer adaptively reduces its fetch size toward $B_{min}$. By pacing consumption and retaining a micro-batch reservoir, ERS extends the step duration, granting Rollout time to accumulate a larger batch for the next iteration.

\vspace{0.45ex}
\noindent\textbf{Seamless Scaling Operations.}
Traditional elastic frameworks suspend training to migrate model weights across nodes. \sys eliminates this latency by exploiting the mathematical properties of on-policy RL.

\vspace{0.5ex}
\noindent\textbf{Cache-Free Task Migration via CTB Coordination.} Iterative model weight updates render historical KV caches mathematically incompatible with the new policy. Thus, caches across Rollout ranks must be flushed at the post-update boundary. Piggybacking on this invalidation cycle, task migration becomes cache-free. \sys bypasses physical KV cache transfers. When ERS closes a Rollout rank, the CTB scheduler automatically reassigns the active tasks on it to the remaining active ranks. Upon resumption, these tasks perform a fresh prefill using the synchronized new version of weights on their new Rollout ranks.

\vspace{0.5ex}
\noindent\textbf{Latency-Hiding Role Switching.} To ensure dynamic resource reallocation does not block the training loop, \sys overlaps PCIe weight-swapping operations with the \texttt{sync\_params} broadcast. When scaling down a Reference rank (to reassign it to Rollout), ERS offloads the reference model to the CPU while waiting for the Actor backward pass to complete. The reactivated Rollout rank then joins the \texttt{sync\_params} broadcast to load the Actor. Conversely, when scaling down a Rollout rank, it discards the stale Rollout model and skips the weight download. Instead, it utilizes the synchronization time window to load the Reference model concurrently with the other ranks' broadcast. Through this operational alignment, \sys adjusts its functional GPU distribution without adding latency to the critical path.

\section{Evaluation}\label{sec:eval}
We implement \sys with $\sim$16000 lines of Python code, supporting mainstream training~(\eg, Megatron~\cite{megatron}, PyTorch FSDP~\cite{fsdp}) and serving~(vLLM~\cite{vllm}, SGLang~\cite{sglang}) frameworks. We disclose the implementation details in Appendix~\ref{sec:appx:impl_details}. In this section, we use vLLM as the rollout backend and Megatron as the training backend.

We evaluate \sys across text-only and multi-modal tasks, model sizes, and readiness regimes. Our key findings are:
\begin{icompact}
\item \sys achieves over 5.6$\times$ RL training throughput on text-only tasks and reduces training time by 51.1\% to reach similar task performance.
\item For multi-modal tasks, \sys improves RL throughput by over 33\% and reduces training time by 62.2\% for similar task performance.
\item CTB improves KV cache hit rate by 1.58$\times$ and generation throughput by 1.15$\times$ by mitigating \textbf{C1}.
\item \rap reduces per-step training time by up to 44.3\% by selecting the proper ref-actor execution mode for different readiness patterns~(\textbf{C2}).
\item ERS uses readiness signals to reallocate GPUs between rollout and training, reducing total waiting time by up to 77.6\%~(\textbf{C3}).
\end{icompact}

\subsection{Methodology}
\label{sec:eval_methodology}

\noindent\textbf{Testbed Configuration.}
We evaluate \sys on a physically disaggregated cluster that completely separates the RL training backend from the environment simulation frontend. The training testbed consists of four compute nodes. Each node is equipped with 8$\times$ NVIDIA H100 GPUs interconnected via NVLink, a 64-core Intel Xeon CPU, and 1.5 TB of RAM memory. To support high-throughput distributed checkpointing and rapid parameter synchronization, a shared JuiceFS-backed NFS is deployed across the training cluster. 

\noindent\textbf{Workloads and Tasks.}
To comprehensively evaluate \sys across diverse agentic scenarios, we select a mixture of standard text-based and complex multimodal tasks. All the tasks simulate varying lengths of trajectories, and are widely adopted by researchers~\cite{agentbench, visualagentbench}. For text-based workloads, we employ a hybrid task suite comprising WebShop~\cite{webshop} and AlfWorld~\cite{alfworld}.  For multi-modal workloads, we utilize OSWorld~\cite{osworld} and ScienceBoard~\cite{sciboard}. They require the agent to process high-resolution screenshots and interact with graphical user interfaces (GUIs), heavily stressing the prefill and context-carriage capacities of Rollout. In the meantime, multi-modal tasks allow a larger allowed turns of interactions.  We disclose the different behaviors of these workloads in Table~\ref{tab:task_stats} of Appendix~\ref{sec:appx:tasks}. We conduct an evaluation every 20 training steps to evaluate the training performance.

\noindent\textbf{Disaggregated Environment.}
We utilize a dedicated bare-metal Kubernetes cluster provisioned with 1024 CPU cores to run environments. This cluster exposes APIs to orchestrate environment lifecycles, enabling the Rollout workers to start, interact with, and close external simulation instances without blocking the model execution threads.

\noindent\textbf{Models.}
We evaluate the system using a wide spectrum of state-of-the-art open-weights models to demonstrate its architectural generality. For text-only tasks, we employ the Qwen-2.5 series (7B and 14B,~\cite{qwen25}). For multimodal tasks, we utilize advanced multi-modal language models, Qwen-3-VL (4B, ~\cite{qwen3vl}), Qwen-3.5 (9B, ~\cite{qwen35}). These models have diverse architectures. Qwen-3.5 also employs linear attention and multi-token prediction~\cite{mtp}. This diverse selection of model scales and architectures allows us to analyze the system's performance and memory management efficiency across different compute-to-memory-bandwidth regimes.

\noindent\textbf{Baselines.}
To evaluate \sys against the relevant design points, we compare it with three RL training frameworks:
\begin{icompact}
    \item \textit{VeRL}~\cite{verl}: A highly optimized synchronous RL framework that enforces strict phase barriers between generation and training. It serves as our primary synchronous baseline to demonstrate the massive idle overheads inherent in coupled architectures.
    \item \textit{AReaL}~\cite{areal}: A foundational asynchronous RL framework that physically decouples the Rollout and Trainer workers. It explicitly utilizes the active partial rollout mechanism introduced by APRIL~\cite{april} to manage off-policy staleness and improve concurrency, representing standard request-level asynchronous scheduling.
    \item \textit{StreamRL}~\cite{streamrl}: An advanced asynchronous framework featuring stream generation support. By streaming trajectories directly to the Trainer without waiting for the entire global batch, it overlaps pipeline execution and represents the current state-of-the-art in asynchronous RL scheduling. For fair comparison, StreamRL always utilize the maximum-allowed GPUs.
\end{icompact}

All frameworks use the same GPU budget, task stream, model checkpoints, staleness bound, global batch configuration, and rollout/training backends whenever the framework supports them. For fixed-partition asynchronous baselines, we sweep the rollout GPU ratio over the same candidate set used in Figure~\ref{fig:resource_util} and report the best-performing setting for each workload. \sys starts from the same candidate ratio, but ERS may reassign ranks during execution. We choose not to include Seer~\cite{seer} as a baseline because its synchronous architecture is represented by VeRL, and its suffix decoding optimization increases rollout time on our multi-turn workloads due to draft-verification overheads, as discussed in~\cref{sec:discuss}.

\noindent\textbf{Metrics.} We use training throughput as the primary system metric: the number of generated tokens that are eventually consumed by the Trainer per second. This excludes stale or discarded rollout tokens and therefore measures useful progress for on-policy RL. We also report task performance, KV cache hit rate, generation throughput, RWT, and TWT to explain where the end-to-end gains come from.
In other words, training throughput is our operational measure of RL training goodput.

\subsection{End-to-End Performance on Text-Only Models}\label{sec:eval:text}
We run \sys and the baselines for 100 training steps. Unless otherwise stated, fixed-partition asynchronous systems use the best static rollout ratio from our sweep; on this workload, that ratio is 0.25.

\begin{figure}
    \centering
    \includegraphics[width=0.85\linewidth]{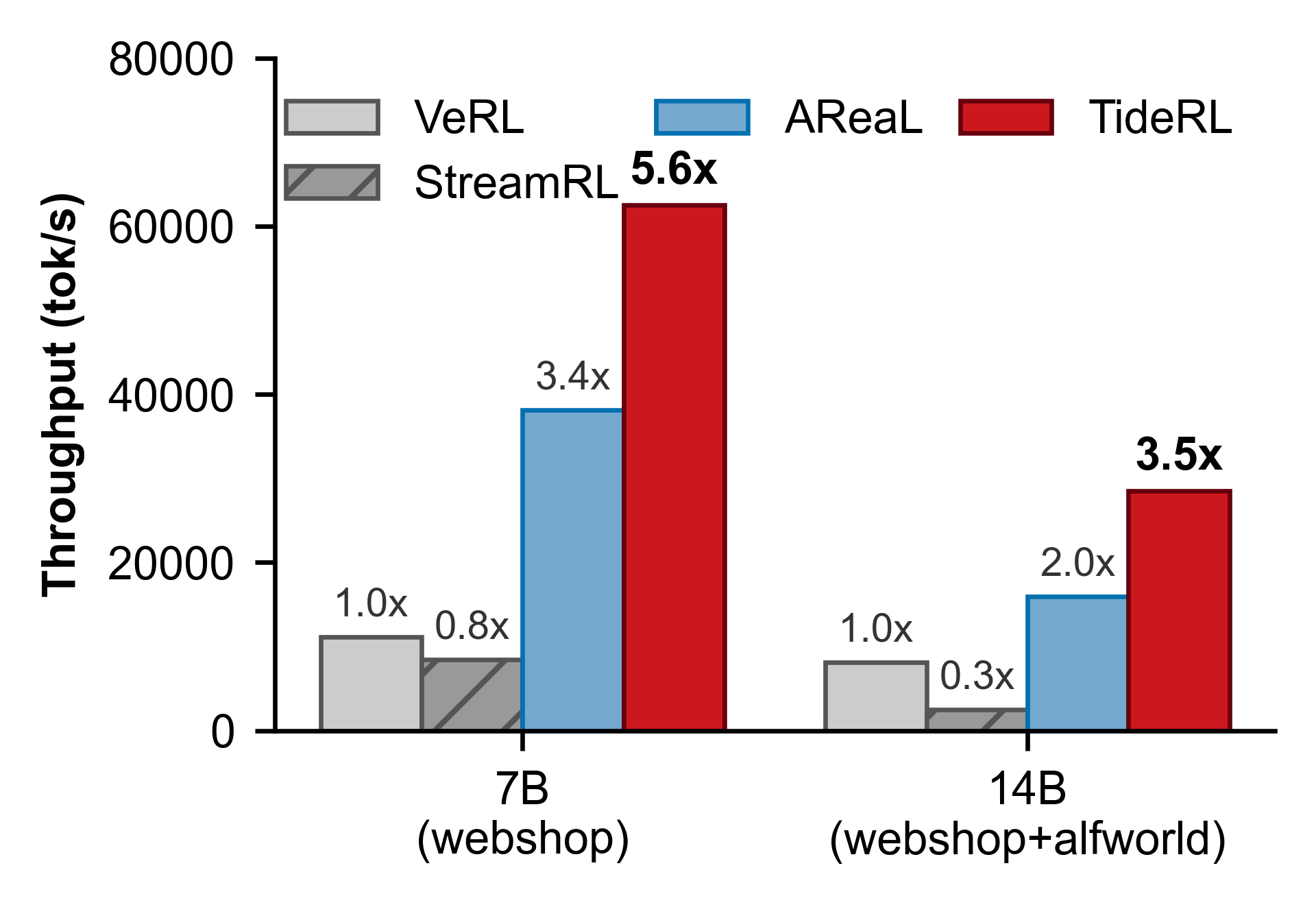}
    \caption{\sys achieves the highest throughput for different models on text-only tasks across four frameworks.}
    \label{fig:throughput_text}
\end{figure}
\noindent\textbf{Throughput.} Figure~\ref{fig:throughput_text} reports training throughput across models and frameworks.

Compared with the synchronous VeRL baseline, \sys achieves a 5.6$\times$ speedup. VeRL is bottlenecked by the slowest trajectories in each global batch: a wrong action can require extra environment turns and substantially extend rollout time. Its colocated synchronous execution also repeatedly swaps Rollout and Trainer states at every step, further reducing useful training throughput.

AReaL improves over VeRL by decoupling generation from training, but its best fixed partition is still a single compromise across different readiness regimes. When evaluation or long-tail tasks make readiness sparse, Trainer stalls; when simple tasks make readiness dense, Rollout stalls behind the Trainer. \sys reacts to these regimes through ERS and the two \rap modes, achieving a 1.8$\times$ throughput improvement over AReaL. \cref{sec:eval:extend} breaks down this effect in the RAS/TPRM plane.

StreamRL reduces global-batch waiting by streaming micro-batches, but on text workloads each DP rank can receive more than 16 micro-batches per step. This bursty stream triggers frequent Ref-Actor model alternation and dominates the saved stall time; the 14B run does not complete within six hours. \sys keeps the streaming benefit while avoiding model thrashing with \rap, yielding over 7$\times$ speedup in this setting.

\begin{figure}
    \centering
    \includegraphics[width=0.8\linewidth]{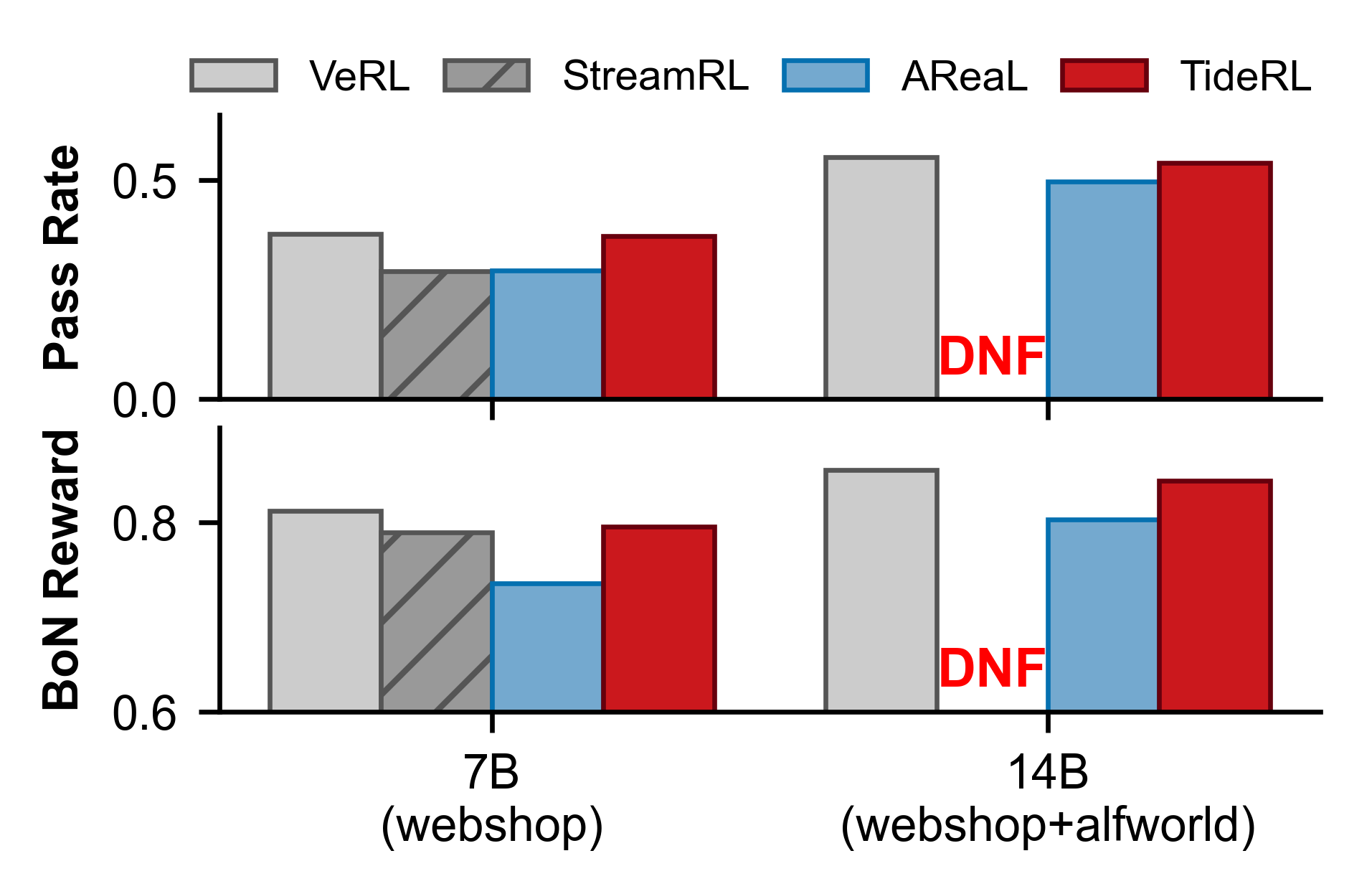}
    \caption{The BoN reward and pass rate for different models on text-only tasks.}
    \label{fig:performance_text}
\end{figure}

\vspace{0.5ex}
\noindent\textbf{Training Performance.} We further exhibit the Best-of-N (BoN,~\cite{bon}) reward and the pass rate after the 100-th training step in Figure~\ref{fig:performance_text}. The strictly synchronized VeRL has all the training data perfectly on policy with zero staleness. As expected, it exhibits the best training performance.

Among the asynchronous RL frameworks, \sys achieves the best task performance and remains close to VeRL, with a BoN reward deficit of 0.01 and a pass-rate deficit of 0.5\%. It reaches this performance using only 48.9\% of VeRL's wall-clock time.

Compared with StreamRL and AReaL, \sys generates and consumes useful trajectories faster, so more of its training data is consumed within the zero-staleness window. The other asynchronous baselines more often fall back to one-step-stale data, which explains why \sys is closer to the synchronous learning curve while retaining much higher throughput.

\subsection{End-to-End Performance on Multi-Modal Models}\label{sec:eval:multi_modal}
We run \sys and the baselines on multi-modal tasks and evaluate both throughput and task performance. Compared with text-only workloads, multi-modal trajectories have longer environment interactions, larger observations, and more variable rollout/training balance, as detailed in Appendix~\ref{sec:appx:tasks}.

\begin{figure}[tb]
    \centering
    \includegraphics[width=0.85\linewidth]{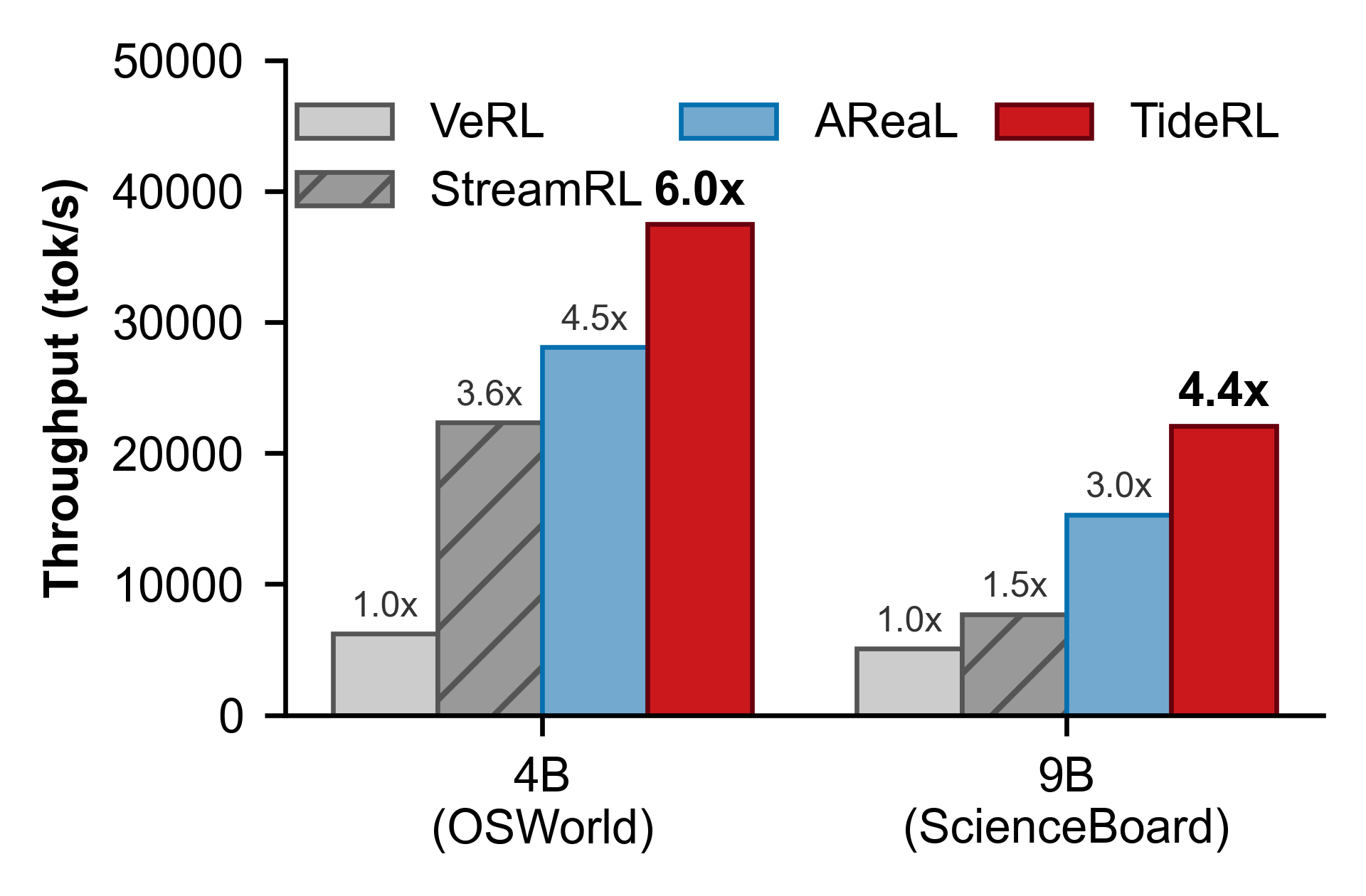}
    \caption{\sys achieves the highest training throughput on multi-modal tasks across four RL frameworks.}
    \label{fig:throughput_mm}
\end{figure}
\vspace{0.5ex}
\noindent\textbf{Throughput.} Figure~\ref{fig:throughput_mm} shows that \sys achieves the highest training throughput on multi-modal tasks.

VeRL suffers more severely in this setting because long GUI interactions and high-variance observations amplify global-batch tail latency.

All asynchronous frameworks improve over VeRL because sparse micro-batch arrivals make overlap more valuable. This regime also reduces StreamRL's model-swapping pressure compared with text tasks. Even after tuning AReaL's fixed ratio to provide sufficient Trainer capacity, \sys achieves 6.02$\times$ throughput over VeRL and more than 1.33$\times$ over the asynchronous baselines. The gain is smaller than in text-only tasks because larger Trainer ranks reduce the number of ranks that ERS can migrate, but readiness-aware scheduling still avoids the worst fixed-partition stalls.

\begin{figure}
    \centering
\includegraphics[width=0.8\linewidth]{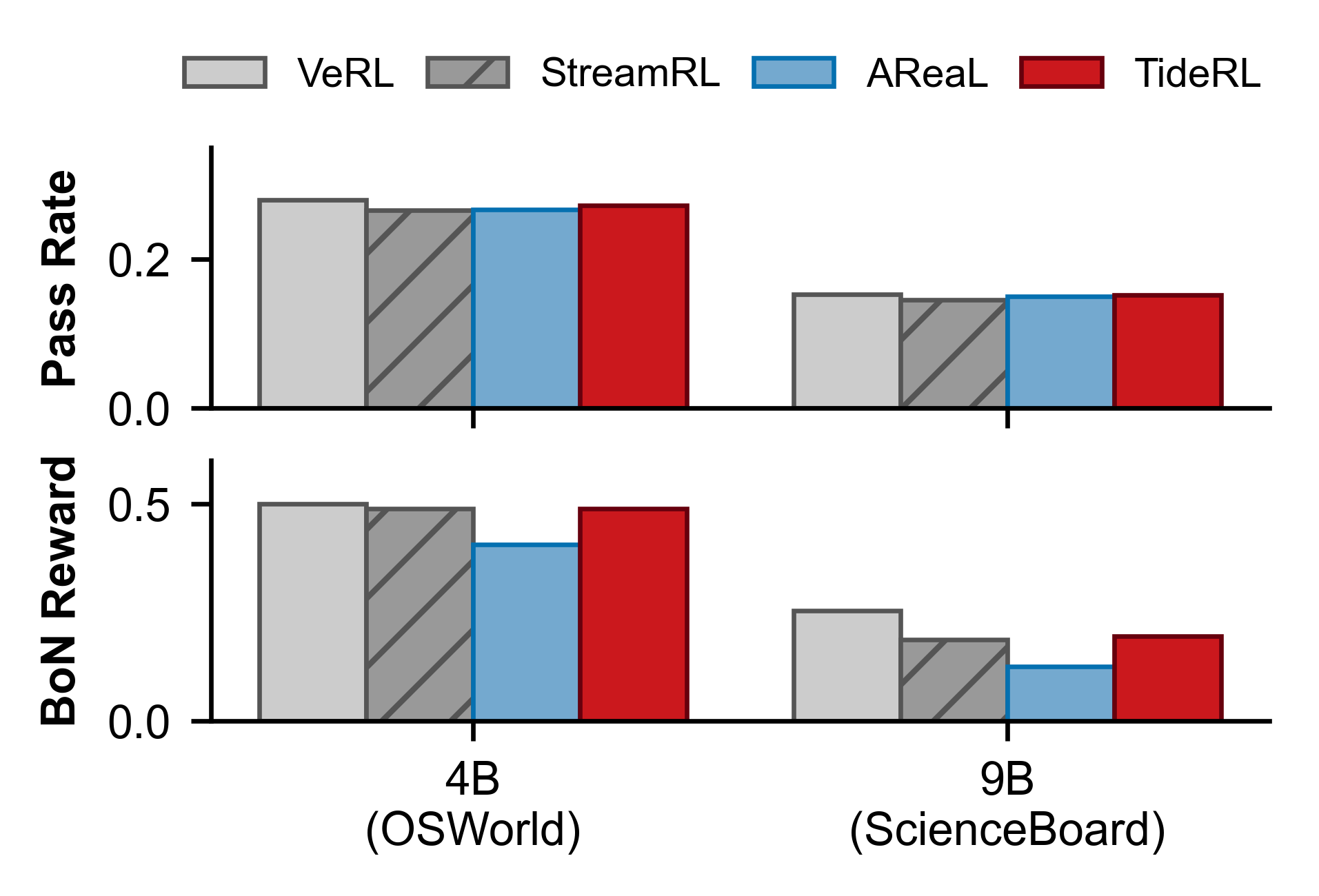}
    \caption{The BoN reward and pass rate for different models on multi-modal tasks.}
    \label{fig:performance_mm}
\end{figure}

\vspace{0.5ex}
\noindent\textbf{Training Performance.} Figure~\ref{fig:performance_mm} reports the BoN reward and pass rate after 40 training steps, where \sys consistently outperforms the three asynchronous baselines. However, due to the high variance in training workloads, ERS generates deliberately conservative scaling plans. Because Trainer scale-up operations are bounded to one rank per step, the Trainer eventually emerges as the long-term pipeline bottleneck, causing the system to naturally gravitate toward a one-step off-policy staleness. While this staleness introduces a slight performance degradation compared to the fully synchronous VeRL, \sys completes the 40 steps in only 37.8\% of the wall-clock time, justifying the trade-off between algorithmic equivalence and training throughput.

\subsection{Improvement Breakdown}\label{sec:eval:breakdown}
To quantify the contribution of each \sys component, Table~\ref{tab:improvement_breakdown} presents an ablation study on OSWorld with Qwen-3-VL 4B. We use StreamRL as the baseline because it already streams rollout data to the Trainer, then add \rap, ERS, and CTB cumulatively. The table reports average throughput over the first 10 steps, including the initial evaluation step used to verify model correctness.

\begin{table}[tb]
\centering
\caption{Throughput improvement breakdown on OSWorld with Qwen-3-VL 4B.}
\small
\label{tab:improvement_breakdown}
\begin{tabular}{lcc}
\toprule
\textbf{Method} & \textbf{Throughput (k token/s)} & \textbf{Improvement}  \\
\midrule
Baseline & 20.2 & -- \\
+\rap & 23.1 & 10.4\%\\
+ERS   & 31.8 & 35.7\% \\
+CTB  & 33.3 & 4.8\% \\
\bottomrule
\end{tabular}
\end{table}

StreamRL's main bottleneck is frequent model swapping across many micro-batches. \rap removes PCIe transfers and GPU context-switching overhead from this path, improving Trainer throughput. ERS further adjusts GPU allocation: after the first step, it scales Rollout down and Trainer up, adding 35.7\% improvement on top of \rap by reducing both RWT and TWT. CTB improves the Rollout side by limiting harmful concurrency and reducing prefix recomputation, especially in the first step where many evaluation tasks run together. This yields a 54.8\% first-step throughput improvement and a 4.8\% overall improvement.

\subsection{Extended Evaluation and Micro-Benchmarks}\label{sec:eval:extend}
We next isolate how each component addresses the bottlenecks in~\cref{sec:system_challenges}.

\begin{figure}
    \centering
    \parbox{0.33\linewidth}{
    \includegraphics[width=0.97\linewidth]{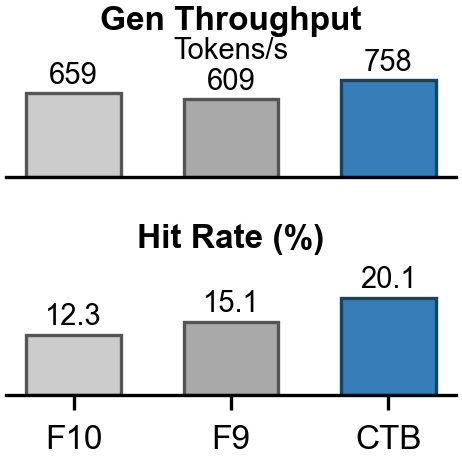}
    \caption{CTB beats F10 and F9 on both metrics.}
    \label{fig:ctb_micro}
    }
    \hfill
    \parbox{0.64\linewidth}{\includegraphics[width=\linewidth]{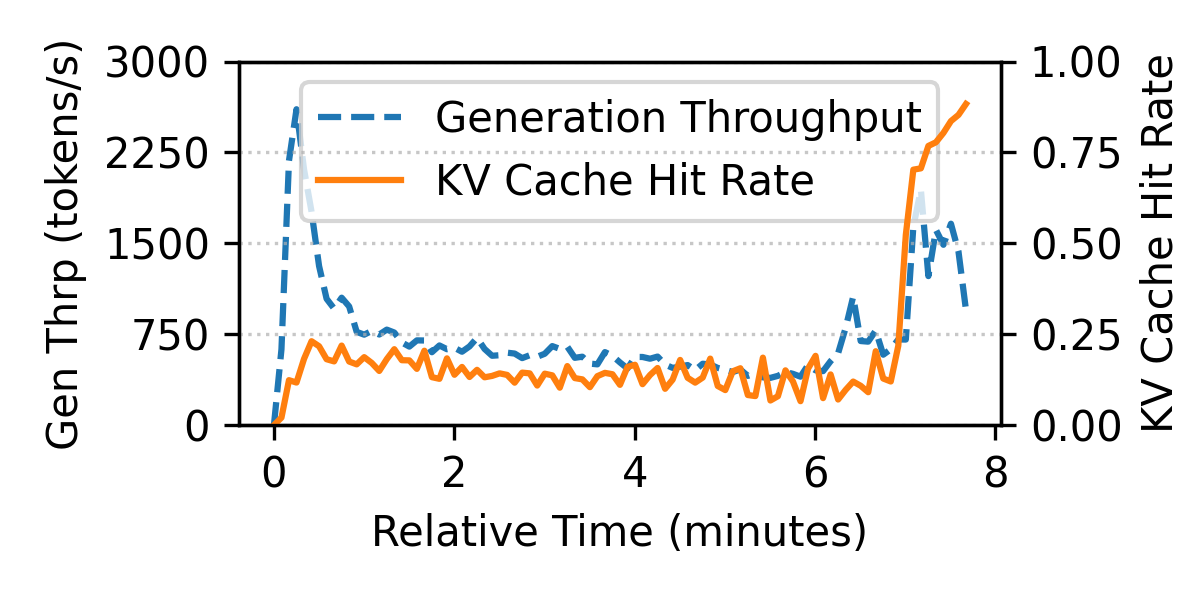}
    \caption{CTB achieves higher throughput and hit rate, in comparison with Figure~\ref{fig:cache_hitrate}.}
    \label{fig:ctb_kv_cache_hit}
    }
\end{figure}

\vspace{0.5ex}
\noindent\textbf{CTB improves KV cache hit rate and rollout throughput.} We perform a step of 1,312 WebShop tasks on one H100. Figure~\ref{fig:ctb_micro} compares CTB with fixed concurrency settings of 1024~(maximum allowed, F10) and 512~(F9). Figure~\ref{fig:ctb_kv_cache_hit} shows that CTB improves cache hit rate by $1.58\times$ and generation throughput by $1.15\times$, reducing step duration by 15.6\%. The vanilla scheduler's frequent preemption drives the cache hit rate as low as 0.9\%, while CTB keeps it above 6.0\%, close to the ideal upper limit. Although hierarchical caching such as SGLang HiCache~\cite{hicache} can expand cache capacity, CTB remains necessary because it avoids excessive PCIe transfers and prefix recomputation. We omit HiCache in our deployment due to initialization overhead and host-memory contention, since the global trajectory buffer and offloaded Trainer model already exhaust the available CPU RAM.

%\vspace{0.5ex}
% \noindent\textbf{CTB priority rules reduce step duration.} Table~\ref{tab:ctb_priority_ablation} isolates CTB's scheduling hierarchy by starting from token-budget admission alone and adding priority rules cumulatively from highest to lowest priority. Step duration decreases as CTB protects the tasks that are most likely to block the next policy update or waste expensive prefix state. Evaluation priority prevents incomplete evaluation trajectories from delaying the weight-synchronization boundary. GRPO group-completion priority lets the Trainer form valid advantage groups earlier. Active-state and context-length protection preserve in-flight computation and long KV prefixes, reducing repeated prefill. Worker affinity further improves resumption by placing paused tasks on ranks that are likely to retain useful shared prefixes. This ablation explains why CTB improves both KV hit rate and rollout throughput in Figure~\ref{fig:ctb_kv_cache_hit}: the gain comes from preserving the right task state, not merely from lowering concurrency.

% TODO: Add Table~\ref{tab:ctb_priority_ablation}.
% Suggested columns: CTB Rule Set, Step Duration, KV Hit Rate, Prefix Recomputed Tokens.
% Suggested rows: Base token-budget admission; +Eval priority; +GRPO group completion;
% +active-state protection; +context-length protection; +worker affinity.

% \vspace{0.5ex}
% \noindent\textbf{RAS/TPRM heatmap explains ERS decisions.} Figure~\ref{fig:ras_tprm_heatmap} compares AReaL and \sys across RAS and TPRM buckets. AReaL keeps a fixed rollout-trainer partition, so different readiness regimes expose different waits: low RAS with long TPRM leaves the Trainer data-starved and increases TWT, while high RAS or short TPRM creates a ready-data backlog that eventually increases RWT. \sys uses the same coordinates as control signals. In low-readiness cells, ERS shifts GPUs to Rollout and selects colocated \rap, avoiding idle dedicated Ref ranks while accelerating future arrivals. In high-readiness cells, ERS shifts capacity to the Trainer and selects decoupled \rap, draining ready micro-batches without repeated model swaps. The heatmap therefore connects the abstract readiness metrics to the measured waiting-time reduction in Figure~\ref{fig:ers_util}: RAS and TPRM identify not only where the bottleneck is, but also which \rap strategy and resource split should remove it.

% TODO: Add Figure~\ref{fig:ras_tprm_heatmap}.
% Suggested layout: two side-by-side heatmaps for AReaL and \sys.
% Axes: RAS bucket and TPRM bucket. Cell value can be step frequency, average step duration,
% or dominant selected \rap mode. If possible, overlay mode decisions for \sys.

\begin{figure}[tb]
\parbox{0.48\linewidth}{
\includegraphics[width=\linewidth]{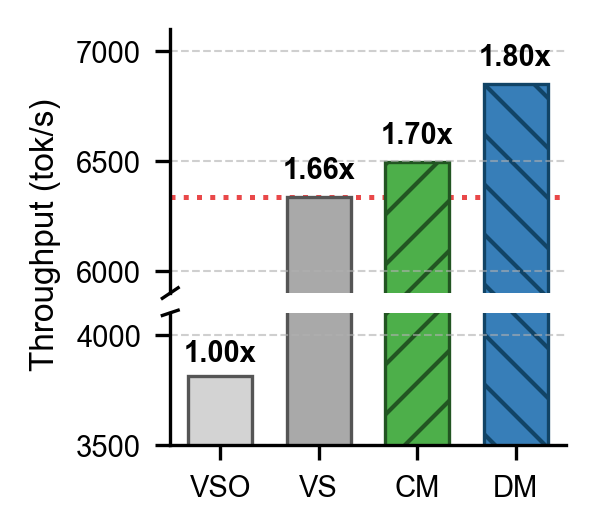}
\caption{\rap~(CM, DM) has higher throughput under ready micro-batch execution.}
\label{fig:rap_util}
}
\hfill
\parbox{0.48\linewidth}{
\includegraphics[width=0.85\linewidth]{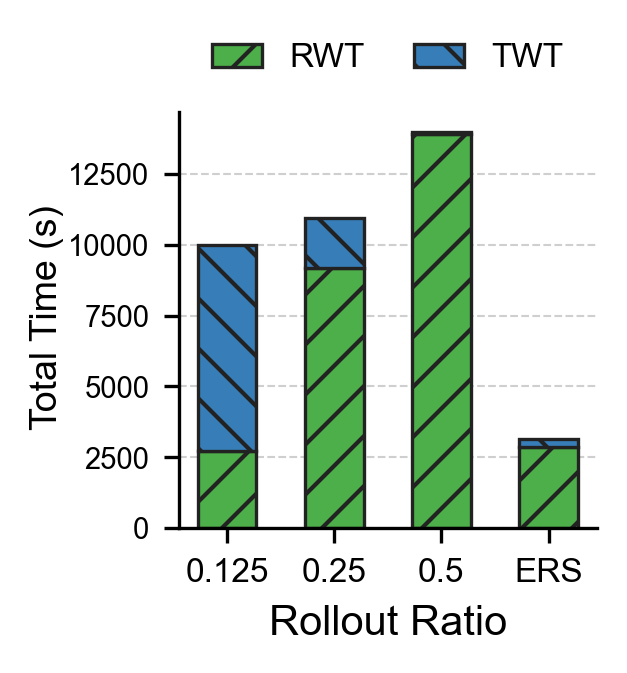}
\caption{ERS relieves RWT and TWT by reacting to readiness imbalance.}
\label{fig:ers_util}}
\end{figure}
\vspace{0.5ex}
\noindent\textbf{\boldrap reduces stall and thrashing.} Figure~\ref{fig:rap_util} shows the overhead for processing eight ready micro-batches on four H100 GPUs with data parallelism only. Each micro-batch has 8,192 tokens. This setup isolates the high-RAS regime where the Trainer begins with enough work to expose model alternation overhead. We compare \rap's colocated mode~(CM) and decoupled mode~(DM) against vanilla streaming with~(VSO) and without~(VS) model offloading. PCIe transfers and GPU context switching introduce significant overhead in the vanilla designs, while \rap reduces overhead by up to 44.3\%. All four methods avoid full-global-batch stalls because they process ready micro-batches immediately.

\vspace{0.5ex}
\noindent\textbf{ERS relieves both Rollout and Trainer waiting.} Figure~\ref{fig:ers_util} reports the sum of RWT and TWT in the first 100 WebShop steps on one 8$\times$NVIDIA H100 node, compared with the fixed rollout ratios in Figure~\ref{fig:resource_util}. Among fixed settings, a rollout ratio of 0.25 gives the best throughput, but it still suffers high RWT in normal steps and high TWT during intermediate evaluation. These phases correspond to alternating large-RAS/short-TPRM and small-RAS/long-TPRM regimes. ERS moves GPUs between Rollout and Trainer and selects the matching \rap mode, reducing total TWT and RWT by 68.6--77.6\%.

\section{Discussion}
\label{sec:discuss}
\begin{table}[tb]
\centering
\caption{Single-instance rollout for Qwen-2.5-7B profiling on WebShop. One step comprises 32 train groups with 8 samples each and 200 eval groups with 4 samples each.}
\label{tab:sd_profiling}
\resizebox{0.95\linewidth}{!}{
\begin{tabular}{lcccc}
\toprule
\textbf{Configuration} & \textbf{Time (s)} & \textbf{Acc. Rate} & \textbf{Acc. Len} & \textbf{Overhead} \\
\midrule
CTB Off (w/o SD) & 548.5 & - & - & - \\
CTB Off (w/ SD)  & 575.4 & 40.1\% & 2.47 & +4.9\% \\
\midrule
CTB On (w/o SD)  & \textbf{460.4} & - & - & - \\
CTB On (w/ SD)   & 495.4 & 43.7\% & 2.83 & +7.6\% \\
\bottomrule
\end{tabular}
}
\end{table}
\noindent\textbf{Incompatibility of Suffix Decoding.}
While suffix decoding (SD, \cite{suffix}) effectively accelerates single-turn inference~\cite{seer}, our profiling shows that it is detrimental to highly variable multi-turn agentic workloads under our model scale. As demonstrated in Table~\ref{tab:sd_profiling}, enabling SD increases total rollout time regardless of CTB status. Complex environmental interactions yield low acceptance rates ($\sim\!40\%$) and short acceptance lengths ($\sim\!2.8$ tokens), so draft-verification overhead outweighs speculative gains. Consequently, SD is excluded from our implementation, though future SD mechanisms could be integrated into \sys when they are beneficial for a workload.

\noindent\textbf{Fault Tolerance and Recovery.} 
In large-scale agentic RL, node failures are inevitable. \sys provides robust fault tolerance without requiring global pipeline restarts. If a Rollout rank fails, the CTB scheduler immediately quarantines the node by halting new task dispatches, while the Trainer seamlessly excludes it from the asynchronous \texttt{sync\_params} broadcast. Furthermore, \sys leverages asynchronous distributed checkpointing~\cite{distcheckpoint, bytecheckpoint} to periodically persist policy weights, ensuring rapid state recovery and minimal interruption to the global training loop.

\noindent\textbf{Algorithmic Effectiveness.} 
In \sys, tokens within a single multi-turn trajectory might be generated by slightly different policy versions as weights update mid-rollout. Consistent with findings in asynchronous RL literature (\eg, APRIL~\cite{april} and AReaL~\cite{areal}), this does not degrade algorithmic convergence. The KL penalty natively regularizes the policy against such version drifts. As demonstrated in \cref{sec:eval}, \sys achieves highly competitive reward growth and learning efficiency compared to strict synchronous baselines.

\noindent\textbf{Limitations and Future Work.} 
While \sys's algorithm design tolerates policy version drift during asynchronous training, evaluation tasks demand an anchored, static model version to ensure metric consistency. Currently, enforcing this strict version synchronization for evaluation can temporarily disrupt the asynchronous pipeline momentum. Future work could address this by integrating a relay weight synchronization mechanism (\eg, Laminar~\cite{laminar}). This would allow the system to maintain decoupled, frozen model snapshots specifically for evaluation tasks without blocking the trainer, provided that the multi-version footprint in host memory can be carefully optimized to be kept with the reference model and large image files in trajectories simultaneously.

\vspace{-1ex}
\section{Related Work}
\label{sec:related}
\noindent\textbf{Distributed RL Systems for LLMs.} 
RL systems for LLMs have primarily optimized GPU utilization under coupled or partially decoupled execution. Synchronous frameworks such as VeRL and RLHFuse~\cite{verl, rlhfuse} colocate generation and training and therefore inherit strict phase barriers. Production frameworks such as SLIME~\cite{slime} and OpenRLHF~\cite{openrlhf} integrate distributed training backends with high-throughput inference engines, but still rely on fixed execution roles during a training job. Recent asynchronous systems overlap rollout and training and reduce long-tail stalls~\cite{streamrl,asyncflow,laminar}; AReaL further bounds policy staleness with active partial rollouts~\cite{areal,april}. These systems expose asynchrony, but they do not jointly decide the ref-actor execution mode and the rollout-trainer GPU split from data-readiness signals. \sys targets this missing control loop: CTB shapes ready-data production, \rap provides two consumption strategies, and ERS schedules both ranks and \rap modes as readiness shifts. Speculative decoding~\cite{seer,specactor}, memory sampling~\cite{infinitesampling}, and hardware-specific dataflows~\cite{mindspeedrl} are orthogonal optimizations that can be integrated when they are beneficial for the workload.

\vspace{0.5ex}
\noindent\textbf{LLM Inference for Agentic Workloads.}
Foundational inference engines optimize throughput via continuous batching~\cite{vllm, orca}, RadixAttention-based prefix sharing~\cite{sglang, cachedattention}, and prefill-decode disaggregation~\cite{distserve, megascale-infer}. Agent-centric serving systems model multi-turn sessions explicitly, adding session-aware scheduling~\cite{agserve, helium}, KV-cache pinning~\cite{continuum}, and congestion control~\cite{concur} to reduce cache evictions during tool-execution pauses. These techniques improve serving throughput, but agentic RL adds algorithmic dependencies absent in serving: GRPO groups must complete before advantages are valid, evaluation tasks must preserve an anchored model version, and rollout staleness must be bounded by training progress. CTB incorporates these dependencies into task admission, pausing, and resumption, so cache residency decisions serve the training loop rather than only request throughput.

\vspace{0.5ex}
\noindent\textbf{Distributed Training Architectures.}
Standard large-scale training relies on 3D parallelism for weight updates~\cite{megatron, gpipe, pipedream, chimera}. DistTrain~\cite{disttrain} disaggregates the visual encoder from the language backbone for multi-modal training. General-purpose elastic frameworks~\cite{pollux, elasticflow, coddl} reallocate resources through co-adaptive batch sizes and throughput modeling, but their scaling operations usually suspend jobs, migrate large model states, or rebuild communication groups. Those costs are too high for agentic RL, where the bottleneck can change across adjacent steps. \sys builds on existing training backends but aligns role switching with the natural post-update synchronization point: stale rollout KV caches are invalidated anyway, and weight movement can be hidden under \texttt{sync\_params}. This makes rank-level elasticity practical inside a fixed-size cluster allocation.

\vspace{-1ex}
\section{Conclusion}
\label{sec:conclusion}
We present \sys, an asynchronous system for multi-turn agentic RL. \sys keeps Rollout efficient with task-level CTB, consumes ready trajectories with the two \rap Ref-Actor strategies, and uses ERS to move GPUs between Rollout and Trainer as bottlenecks shift. On real testbeds, it improves training throughput by over 1.8$\times$ on text-only tasks and over 33\% on multi-modal tasks compared with existing asynchronous systems.

% \clearpage

%\parab{Acknowledgements:} We thank the anonymous SIGCOMM reviewers and our shepherd *** for their constructive feedback and suggestions.
%\parab{Acknowledgements:} We thank the anonymous SIGCOMM reviewers for their constructive feedback and suggestions. This work was supported in part by funding from the National Key R\&D Program of China 2020YFB1807800, the Research Grants Council of Hong Kong (11209520) and CUHK (4055138, 4937007, 4937008, 5501329, 5501517).

%\clearpage
\bibliographystyle{plain}
\bibliography{nsdi27}

\newpage
\appendix
\section*{Appendix}\label{appendix}
\section{Analytical Model for \boldrap Selection}
\label{sec:appendix_rap_cost}

To decide the optimal execution mode in \rap, we use an analytical model to evaluate the total latency to process a global batch of $N$ micro-batches. Let $t_{ref}$ and $t_{act\_f}$ denote the forward pass time for the Ref and Actor models respectively, and $t_{act\_b}$ denote the Actor's backward pass time. Let $t_{swap}$ be the overhead of model swapping. Given the ordered micro-batch ready times $\mathcal{T}=\{a_1,a_2,\dots,a_N\}$ relative to the start of the training step, RAS is $N_0 = |\{i\mid a_i \le 0\}|$, and TPRM is represented by the post-start gaps $\Delta_i = a_i - a_{i-1}$ for $i>N_0+1$.

\noindent\textbf{Cost of Decoupled Streaming Mode.}
In the decoupled mode, models are distributed across separate GPUs, completely eliminating $t_{swap}$. However, idle bubbles are not entirely absent; they exist as \textit{pipeline fill bubbles} during the startup phase. 

As illustrated in the timeline (see Figure ~\ref{fig:rap_pipeline}), multiple Actor ranks initiate their forward passes concurrently. Assuming an Actor group size of $M$, Actor $i$ ($0 \le i < M$) finishes its forward pass at $t \approx t_{act\_f}$. However, the single Ref model processes the reference forward passes sequentially. The \texttt{ref\_log\_probs} for Actor $i$ are only generated at $t \approx (i+1) \cdot t_{ref}$. Consequently, Actor $i$ must wait for a duration of $i \cdot t_{ref}$ (assuming $t_{act\_f} \approx t_{ref}$) before initiating its backward pass $B_i$. 

Beyond this initial pipeline flush overhead, the steady-state throughput is governed by the maximum of each ready gap and the Actor's computation time ($t_{act\_f} + t_{act\_b}$). Micro-batches that are already ready at the step boundary have zero ready gap, so a large $N_0$ immediately amortizes the pipeline fill bubble. The total latency can be approximated as:
$$ C_{dec} \approx \underbrace{\sum_{i=0}^{M-1} (i \cdot t_{ref})}_{\text{Pipeline Fill Bubble}} + \sum_{i=1}^{N} \max(\delta_i, t_{act\_f} + t_{act\_b}) $$
where $\delta_i=0$ for $i \le N_0$, $\delta_{N_0+1}=\max(a_{N_0+1},0)$, and $\delta_i=a_i-a_{i-1}$ otherwise.

\noindent\textbf{Cost of Optimized Colocated Mode.}
In the colocated mode, the dynamic ready-batch aggregation dictates that the aggregation size $k$ is strictly non-deterministic and monotonically adjusts based on real-time task arrivals. Therefore, the execution latency cannot be expressed as a closed-form static formula. Instead, the total time $C_{col}$ is determined algorithmically by simulating the scheduler's progression. 

The scheduler maintains a global clock $T$, initialized to $T = 0$. The evaluation proceeds as follows for the unprocessed micro-batches starting at index $j = 1$:
\begin{enumerate}
    \item \textbf{Wait for Data:} If the current time $T < a_j$, the system idles until the next micro-batch arrives: $T \leftarrow a_j$.
    \item \textbf{Determine Aggregation ($k$):} The scheduler counts all ready micro-batches that have arrived by time $T$. Let $k$ be the number of batches such that their arrival time $a_{j+k-1} \le T$, capped by the maximum memory capacity.
    \item \textbf{Execute Ref Model:} The system incurs a swap overhead, then sequentially processes $k$ reference passes. The clock updates: 
    $$ T \leftarrow T + t_{swap} + k \cdot t_{ref} $$
    \item \textbf{Execute Actor Model:} The system incurs another swap, followed by the Actor's forward and backward passes. The clock updates:
    $$ T \leftarrow T + t_{swap} + k \cdot (t_{act\_f} + t_{act\_b}) $$
    \item \textbf{Advance Index:} The pointer advances ($j \leftarrow j + k$) and the cycle repeats until $j > N$. $C_{col}$ evaluates to the final clock time $T$.
\end{enumerate}

\noindent\textbf{Qualitative Mode Analysis.}
Based on the formulated cost models, the selection between the two execution modes hinges on both $N_0$ and the post-start ready gaps. A large $N_0$ means the Trainer already has a startup burst to process, so the initial pipeline fill bubbles in $C_{dec}$ are rapidly amortized. Short post-start ready gaps then keep the decoupled pipeline saturated. The decoupled mode provides superior throughput in these regimes by strictly eliminating the $t_{swap}$ overhead, making it the optimal choice when maximizing processing speed is the priority and GPU resources are sufficient.

Conversely, when $N_0$ is small and post-start ready gaps are long, data readiness dominates the overall latency. Under these conditions, the colocated mode naturally absorbs the model-swapping penalty within the physical generation delays (Step 1 of the colocated algorithm). Consequently, the colocated mode maintains comparable end-to-end latency to the decoupled pipeline while effectively halving the required GPU footprint, maximizing overall resource efficiency in bottlenecked regimes.

\section{Implementation Details}\label{sec:appx:impl_details}
\subsection{Environment Interfacing and Lifecycle Management}
\sys interacts with external task environments through a standardized set of lightweight RESTful-like APIs. The three primary endpoints are, \texttt{/start}, \texttt{/observation}, and \texttt{/end}. 

In asynchronous agentic RL, tasks may enter a long pause by CTB, causing standard environments to falsely detect inactivity and terminate the session. To solve this, \sys introduces two critical supplementary endpoints: \texttt{/pause} and \texttt{/resume}. These APIs define a protected temporal window. Upon eviction by the CTB scheduler, the \texttt{/pause} signal explicitly halts the environment's internal timeout counters, guaranteeing that long-running sessions and their associated external states (\eg, Docker containers or browser contexts) remain intact until execution resumes.

\subsection{Rollout Engine and CTB Observability}
To execute token-aware Continuous Task Batching (CTB), the system must accurately track sequence lengths across the cluster. \sys implements dual-mode length awareness:
\begin{icompact}
    \item \textbf{Active Reporting:} The user-defined agentic loop proactively reports token consumption and trajectory metadata back to the global scheduler at each interaction turn.
    \item \textbf{Engine Hijacking:} For non-intrusive tracking, \sys dynamically hijacks the scheduling APIs of the underlying Rollout execution engines. This extracts real-time GPU scheduling metadata directly from individual Rollout ranks, avoiding reliance on rough heuristics.
\end{icompact}

Further, \sys natively interfaces with the Prometheus endpoints exposed by SGLang and vLLM. This integration provides granular observability into KV cache utilization, prefill latencies, and decode throughput, serving as foundational metrics for system profiling and performance tuning.

\subsection{Trainer Modifications and Sequence Dispatching}
\sys executes customized data dispatching and micro-batch scheduling to accommodate the asynchronous nature of agentic trajectories. We also include trainer metrics, such as loss to our Prometheus, so that users can easily monitor if it is algorithmically correct, just like SLIME~\cite{slime}.

\begin{icompact}
    \item \textbf{Asynchronous Micro-Batching (Megatron Core):} We modified Megatron Core to decouple forward and backward execution from the strict requirement of a fully assembled global batch. This permits the pipeline to initiate execution on partial micro-batches immediately upon arrival. Notably, we retain the original 1F1B (One-Forward-One-Backward) pipeline schedule without modification. Because \sys does not employ Virtual Pipeline Parallelism (VPP), the absolute pipeline completion time is strictly determined by the arrival and execution of the final micro-batch; altering the schedule interleaving yields no structural latency reduction.
    \item \textbf{Logical DP Grouping (FSDP):} In architectures like PyTorch FSDP, different FSDP ranks within a same DP group mandate collective data fetching and synchronized gradient reduction. To prevent intra-group stragglers, \sys partitions physical DP ranks into \textit{logical DP ranks}. Sequences of highly similar lengths are tightly packed and fed to logical ranks within the same DP group, ensuring symmetric processing times and eliminating synchronization barriers.
\end{icompact}

\section{Detailed Task Specifications}
\label{sec:appx:tasks}

To comprehensively evaluate \sys across diverse agentic scenarios, we detail the interaction constraints and trajectory characteristics for our workloads. These are categorized into text-only and multi-modal tasks based on the nature of their observations and interactions.

\subsection{Text-Only Workloads}
For standard text-based tasks, we enforce a standardized upper bound of 20 interaction turns and a maximum context window of 8192 tokens per trajectory. Table~\ref{tab:task_stats} summarizes the specific interaction statistics for these benchmarks, integrating environment-specific data from systematic evaluations.

\begin{icompact}
    \item \textbf{WebShop} simulates a real-world e-commerce experience with approximately one million products. These tasks average 13 interaction turns and approximately 3000 tokens per trajectory, reflecting high-density HTML-based observations.
    \item \textbf{AlfWorld-WebShop Suite} includes \texttt{AlfWorld} along with the WebShop dataset. AlfWorld is an embodied household agent, ordering agents to accomplish household tasks, such as finding items. We group the two types, and try to make the agent capable of different tasks.
    \item \textbf{Environmental Complexity}: While our evaluation enforces a 20-turn limit for these tasks, inherent complexity varies. For example, \texttt{AlfWorld} and \texttt{WebShop} presents totally different interaction patterns. What's more, as the training goes, we observe that the average turns and the average sequence lengths tend to decrease as the training step increases.
\end{icompact}

\subsection{Multi-Modal Workloads}
Complex multi-modal and long-horizon tasks, such as \texttt{OSWorld} and \texttt{ScienceBoard}, are permitted up to 50 interaction turns to accommodate the increased complexity of GUI-based navigation. To balance visual fidelity with context efficiency, we enforce a sliding window strategy for visual context retention across both environments. Specifically, screenshots are appended sequentially at each turn. Once a trajectory accumulates ten images, the system preemptively prunes the historical visual context, retaining only the five most recent images. This mechanism splits the trajectory and strictly bounds the dynamic image allowance within a limit per step, avoiding overly long trajectories. Images are captured at a resolution of $1280\times720$. In the meantime, it also takes longer for multi-modal tasks to start an environment, which is another reason for them to have a lower throughput.

\begin{icompact}
    \item \textbf{OSWorld} requires agents to process high-resolution screenshots and interact with general-purpose graphical user interfaces (GUIs), such as Microsoft Office. We run the task with 5--9 image allowance.
    \item \textbf{ScienceBoard} is a comprehensive benchmark designed to evaluate multimodal autonomous agents in realistic scientific workflows. These tasks demand that agents navigate specialized scientific software GUIs and process complex domain-specific visual data (\eg, molecular structures or data plots). We run the task with 8--15 image allowance.
\end{icompact}

\begin{table}[h]
\centering
\caption{Task Trajectory Statistics}
\label{tab:task_stats}
\footnotesize
\begin{tabular}{lccc}
\hline
\textbf{Workload} & \textbf{Avg. Rounds} & \textbf{Avg. Tokens} & \textbf{Image Allowance} \\ \hline
\multicolumn{4}{l}{\textit{Text-Only Tasks}} \\
\texttt{WebShop} & 8$\to$5 & 3400$\to$2400 & N/A \\
\texttt{AlfWorld} & 16$\to$11 & 7400 & N/A \\
\hline
\multicolumn{4}{l}{\textit{Multi-Modal Tasks}} \\
\texttt{OSWorld} & 35$\to$15 & 12000 & 5--9 (Sliding) \\
\texttt{SciBoard} & 39$\to$29 & 24000$\to$23000 & 8--15 (Sliding) \\ \hline
\end{tabular}
\end{table}

\end{document}